\documentclass[a4paper, 11pt]{article}

\makeatletter
\renewcommand\paragraph{\@startsection{paragraph}{4}{\z@}%
	{-3.25ex\@plus -1ex \@minus -.2ex}%
	{1.5ex \@plus .2ex}%
	{\normalfont\normalsize\bfseries}}
\makeatother

\usepackage[utf8]{inputenc}
\usepackage{amsmath}
\usepackage{mathtools}
\usepackage[Algorithmus]{algorithm}
\usepackage{algpseudocode}
\MakeRobust{\Call}
\usepackage{graphicx}
\usepackage[colorinlistoftodos]{todonotes}
\usepackage[english]{babel} 
\usepackage[T1]{fontenc}
\usepackage[nonumberlist,acronym,toc]{glossaries}
\usepackage{url}
\usepackage{booktabs}
\usepackage{multirow}
\usepackage{array}

\usepackage{tabulary}
\usepackage{tabularx}
\usepackage{csquotes}
\MakeOuterQuote{"}
\usepackage{longtable}
\usepackage[onehalfspacing]{setspace}

\usepackage{listings}

\usepackage{float}
\usepackage{caption}

\usepackage{pgf}
\usepackage{tikz}
\usepackage{pgfplots}
\usepackage{pgfplotstable} 
\usepackage{tikz-qtree}
\usepgfplotslibrary{external}
\usetikzlibrary{pgfplots.external, arrows, automata,shapes,decorations,decorations.pathreplacing}
\pgfdeclaredecoration{sl}{initial}{
	\state{initial}[width=\pgfdecoratedpathlength-1sp]{
		\pgfmoveto{\pgfpointorigin}
	}
	\state{final}{
		\pgflineto{\pgfpointorigin}
	}
}
\tikzset{external/system call={pdflatex \tikzexternalcheckshellescape -halt-on-error -interaction=batchmode -jobname "\image" "\texsource" && dvips -o "\image".ps "\image".dvi && ps2eps "\image".ps}}
\tikzstyle{e}=[->,shorten >=0.1pt,>=latex]
\tikzstyle{vertex}=[circle, draw, font=\scriptsize, inner sep=0pt, minimum size=20pt]
\tikzstyle{message}=[above,font=\scriptsize, inner sep=13pt]
\tikzstyle{message arrow}=[->,>=stealth,
shorten >=2pt, shorten <=2pt, 
decoration={sl,raise=10pt},decorate]

\pgfplotsset{selectrows/.style 2 args={
		x filter/.code={
			\ifnum\coordindex<#1fi
			\ifnum\coordindex>#2fi
		}
}}

\title{Knowledge representation and diagnostic inference using Bayesian networks in the medical discourse}

\usepackage{authblk}

\author[*]{Sebastian Flügge}
\author[**]{Sandra Zimmer}
\author[***]{Uwe Petersohn}

\affil[*]{\footnotesize{previously TU Dresden, Faculty of Computer Science, Institute for Artificial Intelligence, D-01062 Dresden, Germany}}
\affil[**]{\footnotesize{TU Dresden, Faculty of Computer Science, Institute for Artificial Intelligence, D-01062 Dresden, Germany}}
\affil[***]{\footnotesize{TU Dresden, Faculty of Computer Science, Institute for Artificial Intelligence, D-01062 Dresden, Germany}}

\date{\today}

\usepackage{palatino,graphicx}  
\usepackage[dvips]{epsfig}
\usepackage{psboxit}  
\usepackage{amsthm}
\newtheoremstyle{break}
{\topsep}{\topsep}%
{\itshape}{}%
{\bfseries}{}%
{\newline}{}%
\theoremstyle{break}
\newtheorem{Def}{Definition}
\newtheorem{Bsp}{Example}
\newtheorem{Alg}{Algorithm}
\newtheoremstyle{break}
{\topsep}{\topsep}%
{\itshape}{}%
{\bfseries}{}%
{\newline}{}%
\theoremstyle{break}

\newcommand{\prb}[1]{\mathrm{\mathbf{#1}}}

\usepackage{fancyhdr}
\begin{document}
\def\nummer{\sf TUD-FI19-02 April 2019}
\def\autor{\sf S. Flügge, S. Zimmer \\ U. Petersohn}
\def\institut{\sf Institut f\"ur K\"unstliche Intelligenz}
\def\titel{Knowledge representation and diagnostic inference using Bayesian networks in the medical discourse}

\maketitle
\begin{abstract}
For the diagnostic inference under uncertainty Bayesian networks are investigated. The method is based on an adequate uniform representation of the necessary knowledge. This includes both generic and experience-based specific knowledge, which is stored in a knowledge base. For knowledge processing, a combination of the problem-solving methods of concept-based and case-based reasoning is used. Concept-based reasoning is used for the diagnosis, therapy and medication recommendation and evaluation of generic knowledge. Exceptions in the form of specific patient cases are processed by case-based reasoning~\cite{PGIB09,PZL19}. In addition, the use of Bayesian networks allows to deal with uncertainty, fuzziness and incompleteness. Thus, the valid general concepts can be issued according to their probability. To this end, various inference mechanisms are introduced and subsequently evaluated within the context of a developed prototype. Tests are employed to assess the classification of diagnoses by the network.
\end{abstract}


\section{Introduction}\label{chapter:introduction}

\subsection{Medical background}

Diagnostics in medicine is the assignment of a real case of illness to an ideal image or model of a case of illness. The main parameters of this assignment are anamnesis, symptoms and findings. The physician decides according to the degree of similarity whether a real case corresponds to the ideal image (generic medical knowledge) for a given disease or not.

The recognition of similarity is relatively simple when
\begin{itemize}
	\item the clinical picture is invariant
	\item the disease has a short duration 
	\item the leading symptoms are very specific or
	\item the diagnosis is made rather late or \ always at the same time within the course of the disease
\end{itemize}

For the ideal images of a disease it is true that
\begin{itemize}
	\item in textbooks often all characteristics attributable to the disease describe the ideal image of this disease, although in practice this ideal image is seldom found 
	\item other diseases or characteristics (e.g. age, subjective perception) that may have an effect on the course of the disease remain unnoticed
	\item biological properties such as genetic variability or ontogenetic modifiability are not taken into account and
	\item under certain circumstances the same diseases can have very different courses and consequences.
\end{itemize}	

While ideal images reduce the variability of diseases, in reality diseases are often characterized by heterogeneity. In addition, the often strict deterministic knowledge contained in medical textbooks often does not correspond to medical practice and physicians tend to develop their own individual medical concepts. In general, an older and more specialized physician also has the greater experience and competence. The physician has more knowledge regarding the variability of clinical pictures, their correct assignment, their course or the modification of their course under consideration of therapeutic interventions and depending on disease stages as well as the prediction of the future course. This problem is gaining in importance in connection with new medical paradigms. For example, early intervention by means of detection and early therapy can be used to modify the course of a disease. Such considerations play a role particularly when the diseases in question cannot be cured causally, but can be modified or require long-term treatment. 

Taking these conditions into account, a medical system should be able to support physicians in their complex decisions. 

\subsection{Integration of case-based und concept-based reasoning}\label{CBR}

For a holistic, computer-assisted treatment of patients, both general and experience-based knowledge has to be mapped in a knowledge base. For standard medical cases\footnote{The term standard medical case is often used in the context of evidence-based medicine and describes "normal cases"\,for the treatment of patients. In individual cases, the physician will deviate from this.}, concept-based reasoning, which is based on generic knowledge, is often sufficient. However, since in practice medical experience knowledge is of great importance, case-based reasoning is an indispensable complement. Experiences of medical cases stored in the system are used in the determination of diagnoses and therapies, based on the assumption that similar problems also have similar solutions~\cite{BEKI06}. Due to the obligation for documentation according to § 10 (1) Musterberufsordnung für Ärzte (MBO-Ä) 1997 and from the treatment contract, medical patient information must be legally stored~\cite{MBO97}. If these patient data are recorded in detail, they form a good basis for solving problems with case-based reasoning. In addition, these data occasionally are a necessary compensation for a lack of generic knowledge or for the clarification of generic dependencies~\cite{PZL19,PGIB09,OePe98}.

Concept-based reasoning is used for recommending and evaluating diagnoses, therapies and medications. On the basis of represented generic knowledge the result of concept-based reasoning is the finding of the most specific concepts in the concept base~\cite{PZL19,PGIB09}. This is based on concepts which usually represent generalizations of cases. Based on information about anamnesis, findings, symptoms and risk factors as well as applicable therapeutic measures, diagnoses, findings and therapy recommendations can be derived and proposed to the physician using concept-based reasoning. 

For the solution of medical problems on the basis of specific instantiated knowledge case-based reasoning can be used. It relies on the assumption that similar problem situations also require similar solutions. Within this method, human behavior is imitated. If a new problem arises, comparable situations in the past are searched for. The experience is applied to the current problem completely, partially or with certain adaptations.The prerequisite for case-based reasoning is the existence of specific experiences in the form of cases that are organized in a case base. 

In addition, medical knowledge is strongly characterized by fuzziness, uncertainty and incompleteness. The main reasons for this are the complexity of medicine and the individuality of patients. In their daily clinical work, physicians often have to deal with uncertainties, too. Findings and clinical data may not be completely available at the time of a medically necessary decision. Furthermore, medical knowledge is constantly renewed and physicians have difficulties in keeping their personal knowledge up-to-date in a systematical manner ~\cite{SpSp08}. Through the use of case-based reasoning, uncertain knowledge in the form of patient cases that have been solved previously can provide a supplement. 

For complete problem solving, case-based reasoning and concept-based reasoning are employed for the processing of case-based specific knowledge and for generic knowledge processing, respectively. An integral component of concept-based reasoning are Bayesian networks, which can also process fuzzy concepts. The handling of this uncertainty using Bayesian networks is the subject of this publication.


\subsection{State of the art}


There are different methods for dealing with decisions under uncertainty. The most important approaches are:

\begin{itemize}
	\item \textbf{Bayesian networks:}  \\
	The values of the joint distribution are calculated situation-related as a function of random variables in a subject area. The concept of conditional independence of events is pursued.
	\item \textbf{Dempster-Shafer theory (evidence theory):}  \\
	The quantitative evaluation of the uncertainties is based on subjective assessments of the reliability of expert statements (measures of belief).
	\item \textbf{Fuzzy logic:}  \\
	Each element is assigned a value by means of a membership function, which represents the degree to which it belongs to a fuzzy set. Predicates are thus assigned to fuzzy sets or relations as so-called possibility distributions, which subjectively assess the possibility of them being true.
\end{itemize}

In connection with support for medical decisions, Bayesian networks are preferred because medical knowledge can be mapped in a structured way and the a-priori probabilities often presented in medical systems can be used efficiently for problem solving. 

In general, uncertain knowledge can be represented by a $n$-dimensional random variable $X_1 \dots X_n$, whose values can be interdependent. This forms the basis for the construction of high-dimensional probability distributions using Bayesian networks. These are based on local probabilistic rules that take into account statistical dependencies between characteristics and are thus closely oriented to a causal modelling of the world. Therefore, Bayesian networks can be used to model complex problems with a relatively small number of conditional probabilities~\cite{SpSp05}. 

The aim now is to develop methods for integrating Bayesian networks into diagnostic systems. The following requirements are to be met by the prototype agent implemented for this purpose:

\begin{enumerate}
	\item The definition of the knowledge base should take place externally in the form of an abstract file format, since a representation in source code has proven to be difficult to handle for the modelling of larger networks.
	\item Various inference packages are to be tested, whereby easy integration into existing systems is to be ensured during implementation.
	\item The time complexity for the calculation should be manageable.\footnote{For a statistically Bayesian-optimal decision the probabilities must be known. However, the number of required probabilities can increase so far that with a large number of values the calculations are not controllable due to time complexity.}
\end{enumerate}

For the implementation, various inference algorithms for Bayesian networks were investigated. The inference package \textbf{ebay/bayesian\_networks} is an implementation of the junction tree algorithm. It was developed by eBay, is implemented in Python and is available to the public under an open source license.\footnote{There are also a number of other implementations for Bayesian networks, such as Analytica~\cite{Analytica17}, BayesiaLab~\cite{Bayesia17}, Bayes Server~\cite{Bayesserver17}, JavaBayes~\cite{Javabayes17} or GeNIe/SMILE~\cite{Genie17}.} A detailed description of its functionality can be found in~\cite{beleg} and~\cite{ebaybbn}.

Infer.NET is a development of Microsoft Research, which can be used for inference on different probabilistic graphical models. Therefore it can also be employed to model Bayesian networks. The representation is done by C\# code, which uses the classes of the Infer.NET library\footnote{Since 2018 the source code, which can be found under the MIT license on Github, is available in addition to the compiled library.}. This model is translated by Infer.NET into optimized C\# code on which repeated queries can be made (section~\ref{chapter:implementation:sec:infer.net_agent}). Infer.NET implements three inference algorithms~\cite{InferNET18}:  

\begin{enumerate}
	\item \textbf{Expectation propagation} (EP) is an extension of \textbf{loopy belief propagation} (LBP, section~\ref{sec:Loopy-Belief-Propagation}) which also supports continuous normally distributed variables. Here EP is treated like an implementation of LBP, since only discrete networks are relevant for the discussed discourse area. 
	\item The reference algorithm of Infer.NET is \textbf{variational message passing} (VMP, section~\ref{chapter:theory:sec:inference:subsec:VMP}), which was developed for inference on complex mixed networks of continuous and discrete variables. The inferential performance of the two algorithms on the headache network (figure~\ref{fig:full_network}) is examined in section~\ref{chapter:implementation:sec:test}. 
	\item The third algorithm available is \textbf{Gibbs sampling}, which was not examined any further due to the high computational effort needed to achieve a sufficient accuracy.
\end{enumerate}

The following table gives an overview of the properties of the investigated inference algorithms.

\begin{table}[H]
	\footnotesize
	\centering
	\small
	{\renewcommand{\arraystretch}{1.3}
		\begin{tabulary}{1.0\textwidth\tymin=50pt}{CCCCCC}
			\toprule
			& Deterministic? & Exact result? & Always converging? & Efficient? & Available in Infer.NET? \\ \toprule
			Loopy belief propagation     & Yes   & In single connected networks only & No, but sufficient for many applications & Yes (for discrete networks) & Yes, with expectation propagation\\ \midrule 
			Junction tree belief propagation & Yes   & Yes & Yes & Yes (for discrete networks that are not too complex) & No \\ \midrule 
			Variational message passing  & Yes   & No & Yes & Yes & Yes \\ \bottomrule
	\end{tabulary}}
	\caption{Overview of the inference algorithms for Bayesian networks considered in this report~\cite{InferNET18}.}\label{uebersicht_inferenzalgorithmen}
\end{table}

\section{Representation and Inference}\label{chapter:theory:sec:bn_definition}

Bayesian networks construct a high-dimensional probability distribution based on local probabilistic rules. A Bayesian network is represented as a directed acyclic graph (DAG) in which nodes are random variables and edges describe conditional dependencies between these variables. The (directed) edges represent the causal influence of one node (parent node) on another node.  The influence of the parent node on the corresponding local random variable is assigned to each node in the network by a table of conditional probabilities. Essential principles of Bayesian networks can be found in~\cite{russel}.

\subsection{Definition of a Bayesian network}\label{chapter:theory:sec:bn_definition}
\subsubsection{Definition of the structure}\label{chapter:theory:sec:bn_definition:subsec:structure}

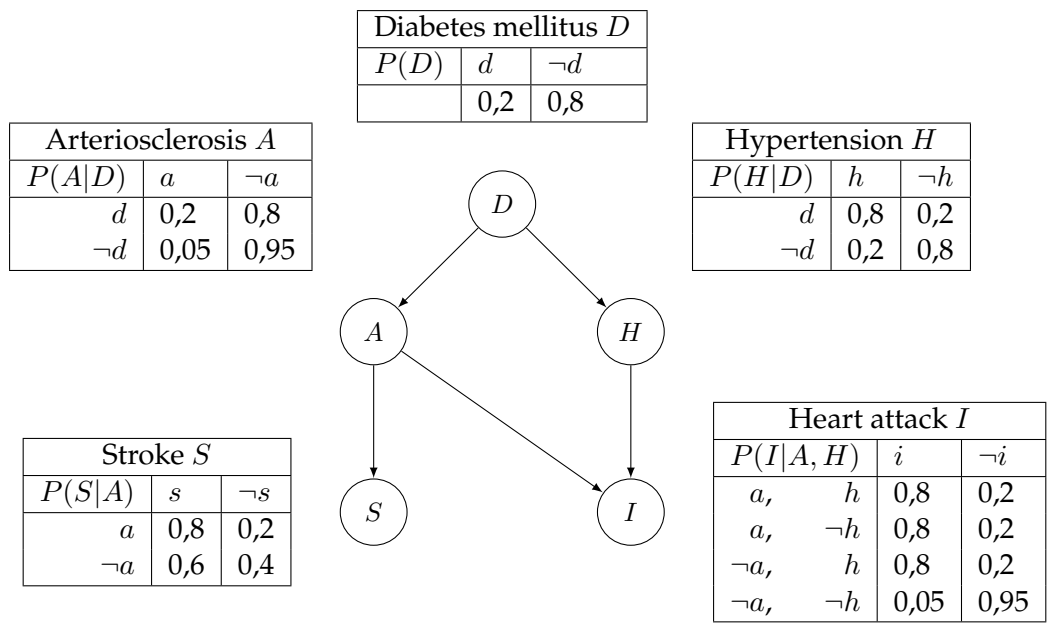
\begin{figure}[H]
	\centering
	\scalebox{1.0}
	{\renewcommand{\arraystretch}{1.0}
		\begin{tikzpicture}[every label/.style={font=\small}]
		\tikzstyle{vertex}=[circle, draw, font=\small, inner sep=0pt, minimum size=25pt]
		\tikzstyle{phantom}=[vertex,draw=none]
		
		\node[vertex] (D) at (0,0) {$D$};
		\node[above=0.5 of D] (DA){
			\begin{tabular}{|c|l|l|}
			\hline
			\multicolumn{3}{|c|}{Diabetes mellitus $D$} \\ \hline
			$P(D)$ & $d$ & $\neg d$ \\ \hline
			& 0,2 & 0,8 \\ \hline
			\end{tabular}
		};
		\node[vertex, below left=1.5 of D] (A) {$A$};
		\node[above left=0.5 of A] (AA) {
			\begin{tabular}{|r|l|l|}
			\hline
			\multicolumn{3}{|c|}{Arteriosclerosis $A$} \\ \hline
			$P(A|D)$ & $a$  & $\neg a$ \\ \hline
			$d$      & 0,2  & 0,8      \\ 
			$\neg d$ & 0,05 & 0,95     \\ \hline
			\end{tabular}
		};
		\node[vertex, below right=1.5 of D] (H) {$H$};
		\node[above right=0.5 of H] (HA) {
			\begin{tabular}{|r|l|l|}
			\hline
			\multicolumn{3}{|c|}{Hypertension $H$} \\ \hline
			$P(H|D)$ & $h$ & $\neg h$ \\ \hline
			$d$      & 0,8 & 0,2      \\ 
			$\neg d$ & 0,2 & 0,8      \\ \hline
			\end{tabular}
		};
		\node[vertex, below=1.5 of H] (I) {$I$};
		\node[right=0.5 of I] (IA) {
			\begin{tabular}{|r r|l|l|}
			\hline
			\multicolumn{4}{|c|}{Heart attack $I$} \\ \hline
			\multicolumn{2}{|c|}{$P(I|A,H)$}
			& $i$  & $\neg i$ \\ \hline
			$a$,     & $h$     & 0,8  & 0,2      \\ 
			$a$,     &$\neg h$ & 0,8  & 0,2      \\
			$\neg a$,&$ h$     & 0,8  & 0,2      \\ 
			$\neg a$,&$\neg h$ & 0,05 & 0,95     \\ \hline
			\end{tabular}
		};
		\node[vertex, below=1.5 of A] (S) {$S$};
		\node[left=0.5 of S] (SA) {
			\begin{tabular}{|r|l|l|}
			\hline
			\multicolumn{3}{|c|}{Stroke $S$} \\ \hline
			$P(S|A)$ & $s$ & $\neg s$ \\ \hline
			$a$      & 0,8 & 0,2      \\ 
			$\neg a$ & 0,6 & 0,4      \\ \hline
			\end{tabular}
		};
		\path
		(D) edge[e] (A)
		(D) edge[e] (H)
		(A) edge[e] (I)
		(H) edge[e] (I)
		(A) edge[e] (S)
		;
		\end{tikzpicture}}
	\caption{Example of a simple Bayesian network used in medicine.}\label{Karzinomnetz}
\end{figure}

A Bayesian network is a probabilistic graphical model (PGM) used to map uncertain knowledge in a closed domain. It consists of a graph and local probability distributions (figure~\ref{Karzinomnetz}). The following properties apply:

\begin{itemize}
	\item Each node represents a random variable $X$ with a range of values that can be discrete or continuous.\footnote{In medical diagnostics, most nodes are only binary.} With $\mathrm{\mathbf{\Pi}}_X$ we denote the set of parent nodes, and with ${\pi}_X$ their instantiations.
	\item The relations between the nodes of the graph are \textbf{directed} and the graph is \textbf{cycle free} (DAG).
	\item Each edge describes a parent-child relationship between two nodes. The source node is called \textbf{parent node}, the destination node is called \textbf{child node}.
	\item Each node contains a \textbf{local probability distribution}. For child nodes, this probability distribution is a conditional probability distribution $\mathrm{\mathbf{P}}(X|\mathrm{\mathbf{\Pi}}_{X})$ that quantifies the effect of the parent nodes $\mathrm{\mathbf{\Pi}}_{X}$ on the node $X$. For nodes with $\mathrm{\mathbf{\Pi}}_X = \emptyset$ unconditional a priori distributions $\mathrm{\mathbf{P}}(X)$ apply.
\end{itemize}

The following figure~\ref{Karzinomnetz} shows an example of a Bayesian network used in medicine for calculating the probability of a stroke and heart attack. Diabetes mellitus $D$ may result in arteriosclerosis (vascular calcification) $A$ and hypertension (high blood pressure) $H$, which in turn may cause the patient to have a heart attack $I$. In addition, arteriosclerosis can lead to a stroke $S$. The probability ratings shown in the figure are fictious. 

Depending on the condition of the network, the probability can be determined a priori or \mbox{a posteriori}. For nodes without parents, the probability values that the random variables can take must be specified a priori, e.g. $P(A=i)\forall i$. In figure~\ref{Karzinomnetz} this is the case with the node \textit{diabetes mellitus}. Here, probabilities with the boolean values \textit{true} or \textit{false} are typically used. For nodes with parent nodes, conditional probabilities such as $P(A=i|B=j,C=k)\forall i,j,k$ are specified. The conditional probabilities are tabulated.\footnote{The definition and quantification of the causal dependencies is generally done by an expert of the considered domain who fills the tables with the conditional probabilities. However, this can be difficult or even impossible, especially if a variable has many parents. In this case, often simplifying structures such as \textit{Noisy-OR} (section~\ref{sec:ICI models:noisy:or}) are employed.} They result from the combination of all parental node values for all values of the target node~\cite{SpSp05,SpSp08}. 

\subsubsection{Representation of a complete common distribution}\label{chapter:theory:sec:bn_definition:subsec:full_distribution}

The complete common distribution, also called \textbf{joint distribution}, contains all the uncertain knowledge about a domain. A Bayesian network can be understood as a representation of such a distribution. Let $P(x_1, \ldots, x_n)$ be a generic entry in the probability table, where $x_1$ to $x_n$ are concrete value assignments to the individual nodes.  The probability of this value assignment can be expressed as the product of the conditional probabilities of the individual nodes as follows:

\begin{align}\label{Verbunddichte}
	P(x_1, \ldots, x_n)=\prod_{i=1}^{n}P(x_i|\mathrm{\mathbf{\pi}}_i).
\end{align}
$\mathrm{\mathbf{\pi}}_i$ are the concrete instances of the parent nodes $\mathrm{\mathbf{\Pi}}_i$ appearing in the distributions $x_1, \ldots, x_n$.

\subsection{Method for constructing a Bayesian network}\label{chapter:theory:sec:bn_construction}

Here a method for creating a Bayesian network shall be shown, so that the resulting network represents the domain as well as possible. This requires conditional independencies, which are implied by~\eqref{Verbunddichte}. With the product rule,~\eqref{Verbunddichte} can also be written as follows:
\begin{align*}
	P(x_1,\ldots, x_n)=P(x_n|x_{n-1},\ldots,x_1)P(x_{n-1},\ldots,x_1).
\end{align*}
Repeated application of the product rule (also called chain rule) results in the following product:
\begin{align*}
	P(x_1,\ldots, x_n)&=P(x_n|x_{n-1},\ldots,x_1)P(x_{n-1}|x_{n-2},\ldots,x_1)\ldots{}P(x_2|x_1)P(x_1)\\
	&=\prod_{i=1}^{n}P(x_i|x_{i-1},\ldots,x_1).
\end{align*}
Compared to~\eqref{Verbunddichte} for every variable $X_i$ of the network and its parent nodes $\mathrm{\mathbf{\Pi}}_i$ it thus holds true that
\begin{align}\label{Nummerierung}
	\mathrm{\mathbf{P}}(X_i|X_{i-1},\ldots,X_1)=\mathrm{\mathbf{P}}(X_i|\mathrm{\mathbf{\Pi}}_i)
\end{align}
provided that $\mathrm{\mathbf{\Pi}}_i\subseteq\{X_{i-1},\ldots, X_1\}$.
From equation~\eqref{Nummerierung} it follows that a Bayesian network only correctly represents a domain if a node $X_i$ with known parents $\mathrm{\mathbf{\Pi}}_i$ is conditionally independent of its remaining predecessors in the node order (section~\ref{sec:conditional_independency}).

The following algorithm generates a network that meets these conditions.

\begin{Alg}[Construction of a Bayesian network]
	Nodes and edges of a Bayesian network can be determined as follows. 
	\begin{enumerate}
		\item \textit{Nodes}: Determine the set of random variables $X_i$ relevant to the domain. In order to achieve a compact network structure, sort it so that causes appear before effects.\footnote{This ensures, among other things, that there are no cycles in the resulting directed graph.} The result is $\{X_1,\dots, X_n\}$.
		\item \textit{Edges}: Execute for $i=1$ to $n$:
		\begin{itemize}
			\item Select from the set $\{X_1, \ldots, X_{i-1}\}$ a minimum subset of parents $\mathrm{\mathbf{\Pi}}_i$ that satisfies the joint density equation.
			\item For each element in $\mathrm{\mathbf{\Pi}}_i$, insert a directed edge from the parent node to the node $X_i$.
			\item Define the conditional probability distribution $\mathrm{\mathbf{P}}(X_i|\mathrm{\mathbf{\Pi}}_i)$.
		\end{itemize}
	\end{enumerate}
\end{Alg}

In other words, $\mathrm{\mathbf{\Pi}}_i$ should contain all nodes from $\{X_1,\dots,X_{i-1}\}$ that directly affect $X_i$.

The algorithm described here guarantees acyclicity, since each node is only connected to nodes that lie before it in the set of nodes sorted by causality.

\subsection{Structural properties}

\subsubsection{Local structuring and compactness}

Bayesian networks belong to the class of locally structured systems. A locally structured system consists of subcomponents, with each component interacting only with a limited number of other components, regardless of the total number of components. Such a system therefore usually exhibits linear instead of exponential complexity growth. 

Applied to Bayesian networks, this means that each random variable is only directly influenced by a limited number of other random variables. Let this number be denoted by the constant $k$ and a network with $n$ boolean variables is considered. Since a maximum of $2^k$ numbers are required to specify the conditional probabilities per variable, only $n2^k$ numbers are required to specify the complete network, while the joint distribution contains $2^n$ numbers. So the larger the number of nodes, the more drastic this difference becomes. Thus, Bayesian networks are a particularly compact representation of a joint distribution.

These assumptions apply to most domains, but there are also issues where each variable is affected by all the others. Modelled as a Bayesian network, this results in a \textit{fully connected} network in which each node is connected to all other nodes by an edge. In this case, the amount of information required to specify the conditional probabilities of the network even exceeds the amount of information required to specify a corresponding joint distribution.\footnote{For a fully connected network, $k=n-1$ applies and $n2^{n-1}$ numbers are required to specify the conditional probabilities for such a network.}

In some domains, dependencies between two variables are very weak. Since the required probabilities to be calculated by the network rarely have to be very accurate, in order to obtain a more compact network these dependencies can be left out of the modelling. \textit{Simply connected} networks, wich are also called polytrees, have a particularly compact structure\footnote{A polytree is a directed graph in which a node can have several root nodes, but in which no node can be reached by another node using different paths. If the directions are removed from the edges, the result is an acyclic graph that can also be displayed as a tree.}. Here for $n$ nodes there are only $n-1$ edges.

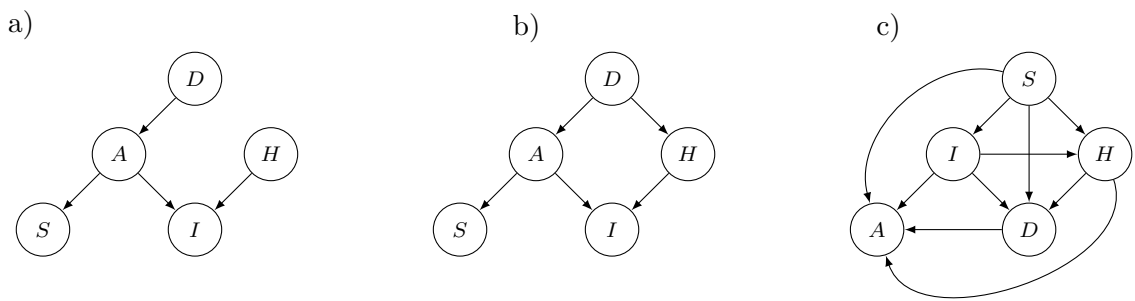
\begin{figure}[H]
	\centering
	\scalebox{1.0}{\begin{tikzpicture}[every label/.style={font=\small}]
		\tikzstyle{vertex}=[circle, draw, font=\scriptsize, inner sep=0pt, minimum size=20pt]
		\tikzstyle{phantom}=[vertex,draw=none]
		\tikzstyle{factor}=[draw, fill, inner sep=0pt, minimum size=15pt]
		\tikzstyle{description}=[decorate,below, inner sep= 30pt, align=left]
		
		\node (A) at (0,0){
			\begin{tikzpicture}
			\node[vertex] (g1_D) at (2,2)
			{$D$};
			\node[vertex] (g1_A) at (1,1) 
			{$A$};
			\node[vertex] (g1_H) at (3,1) 
			{$H$};
			\node[vertex] (g1_S) at (0,0) 
			{$S$};
			\node[vertex] (g1_I) at (2,0) 
			{$I$};
			\path
			(g1_D) edge[e] (g1_A)
			(g1_A) edge[e] (g1_I)
			(g1_A) edge[e] (g1_S)
			(g1_H) edge[e] (g1_I);
			\end{tikzpicture}};
		\draw[decorate] (A.north west) node[above right=-4pt]{a)};
		
		\node[below right=0 and 1.5 of A.north east] (B){
			\begin{tikzpicture}
			\node[vertex] (g1_D) at (2,2)
			{$D$};
			\node[vertex] (g1_A) at (1,1) 
			{$A$};
			\node[vertex] (g1_H) at (3,1) 
			{$H$};
			\node[vertex] (g1_S) at (0,0) 
			{$S$};
			\node[vertex] (g1_I) at (2,0) 
			{$I$};
			\path
			(g1_D) edge[e] (g1_A)
			(g1_D) edge[e] (g1_H)
			(g1_A) edge[e] (g1_I)
			(g1_A) edge[e] (g1_S)
			(g1_H) edge[e] (g1_I);
			\end{tikzpicture}};
		\draw[decorate] (B.north west) node[above right=-4pt and 30pt]{b)};
		
		\node[below right=0 and 1.5 of B.north east] (C){
			\begin{tikzpicture}
			
			\node[vertex] (g1_D) at (2,2)
			{$S$};
			\node[vertex] (g1_A) at (1,1) 
			{$I$};
			\node[vertex] (g1_H) at (3,1) 
			{$H$};
			\node[vertex] (g1_S) at (0,0) 
			{$A$};
			\node[vertex] (g1_I) at (2,0) 
			{$D$};
			\path
			(g1_D) edge[e] (g1_A)
			(g1_D) edge[e] (g1_H)
			(g1_D) edge[e] (g1_I)
			(g1_D) edge[e, bend right=60] (g1_S)
			(g1_A) edge[e] (g1_I)
			(g1_A) edge[e] (g1_S)
			(g1_A) edge[e] (g1_H)
			(g1_H) edge[e] (g1_I)
			(g1_H) edge[e, bend left=90] (g1_S)
			(g1_I) edge[e] (g1_S);
			\end{tikzpicture}};
		\draw[decorate] (C.north west) node[above right=-4pt and 10pt]{c)};
\end{tikzpicture}}
\caption{Different connectivities of Bayesian networks using the example network from figure~\ref{Karzinomnetz}.}\label{fig:local_structure_graphs}
\end{figure}

In figure~\ref{fig:local_structure_graphs} a) a \textit{simply connected} network is shown.\footnote{From the original network from figure~\ref{Karzinomnetz} the causal relationship between $D$ and $H$ was omitted.}\@ b) shows the original network from figure~\ref{Karzinomnetz} as a \textit{multiply connected} network. If, as illustrated in c), a poor node sequence is selected, the worst case scenario is a \textit{fully connected} network.

In a locally structured domain, compactness depends strongly on the node sequence. An unfavourable sequence leads to additional connections in the network, whose probability distributions are very unnatural and therefore difficult to determine. Such a network encodes the conditional independence statements between the variables rather poorly. In general it can be said that a causal sequence of nodes, i.\,e.\ from cause to effect, leads to much more compact networks than a diagnostic sequence. In the medical domain, diagnoses for diseases should therefore be placed before symptoms of diseases. Even experienced physicians prefer to specify probabilities for causal rather than diagnostic rules, which further simplifies the specification of conditional probabilities.

\subsubsection{Conditional Independence}\label{sec:conditional_independency}

As already mentioned in the last section, a graph structure encodes a set of conditional independence statements which are not dependent on the quantification of the network.

A statement of independence in the form \textit{X and Y are independent with known Z} means that the following equation applies to all combinations of the values $x$, $y$ and $z$:
\[\mathrm{\mathbf{P}}(\mathrm{\mathbf{x}}|\mathrm{\mathbf{z}})=\mathrm{\mathbf{P}}(\mathrm{\mathbf{x}}|\mathrm{\mathbf{yz}})\]

In other words: If $z$ is known, the knowledge of $y$ will not influence the value of $x$.

\begin{figure}[H]
	\centering
	\tikzstyle{markov}=[vertex,fill=Snow2]
	\tikzstyle{phantom}=[vertex,draw=none]
	\begin{tikzpicture}[every label/.style={font=\footnotesize}]
	\tikzset{
		between/.style args={#1 and #2}{
			at = ($(#1)!0.5!(#2)$)
		}
	}
	\node[vertex] (X1) at (0,0) [label=above left:$X$]{};
	\node[vertex, right=6 of X1] (Y1) [label=above left:$Y$]{};
	\node[vertex, below=2 of X1] (X3)  {};
	\node[vertex] (Y3) at (X3-|Y1) {};
	\node[vertex, between=Y1 and Y3] (Y2) {};
	\node[vertex, between=X1 and X3] (X2) {};
	\node[vertex, between=Y1 and Y3] (Y2) {};
	\node[vertex, between=X1 and Y1] (Z1) [label=above left:$E$]{$Z$};
	\node[vertex, between=X1 and Z1] (XZ1) {};
	\node[vertex, between=Z1 and Y1] (ZY1) {};
	\node[vertex, between=X2 and Y2] (Z2) {$Z$};
	\node[vertex, between=X2 and Z2] (XZ2) {};
	\node[vertex, between=Z2 and Y2] (ZY2) {};
	\node[vertex, between=X3 and Y3] (Z3) {$Z$};
	\node[vertex, between=X3 and Z3] (XZ3) {};
	\node[vertex, between=Z3 and Y3] (ZY3) {};
	\node[phantom, between=XZ3 and Z3] (P1) {};
	\node[vertex, below=0.5 of P1] (ZC1) {};
	\node[phantom, between=Z3 and ZY3] (P2) {};
	\node[vertex, below=0.5 of P2] (ZC2) {};
	
	\draw ($(X1.north west)+(-0.6,0.6)$)  rectangle ($(X3.south east)+(0.6,-0.6)$);
	\draw[fill=Snow2] ($(Z1.north west)+(-0.6,0.6)$)  rectangle ($(Z2.south east)+(0.6,-0.4)$);
	\draw ($(Y1.north west)+(-0.5,0.5)$)  rectangle ($(Y3.south east)+(0.5,-0.5)$);
	\node[vertex,fill=white] (a) at (Z1) [label=above left:$E$]{$Z$};
	\node[vertex,fill=white] (b) at (Z2) {$Z$};
	
	\path
	(X1) edge (XZ1)
	(XZ1) edge[e] (Z1)
	(Z1) edge[e] (ZY1)
	(ZY1) edge (Y1)
	(X2) edge (XZ2)
	(XZ2) edge[e,<-] (Z2)
	(Z2) edge[e] (ZY2)
	(ZY2) edge (Y2)
	(X3) edge (XZ3)
	(XZ3) edge[e] (Z3)
	(Z3) edge[e,<-] (ZY3)
	(ZY3) edge (Y3)
	(Z3) edge[e] (ZC1)
	(Z3) edge[e] (ZC2)
	;
	\end{tikzpicture}
	\caption{d-separability~\cite{RuNo95}.}\label{fig:d-separation}
\end{figure}
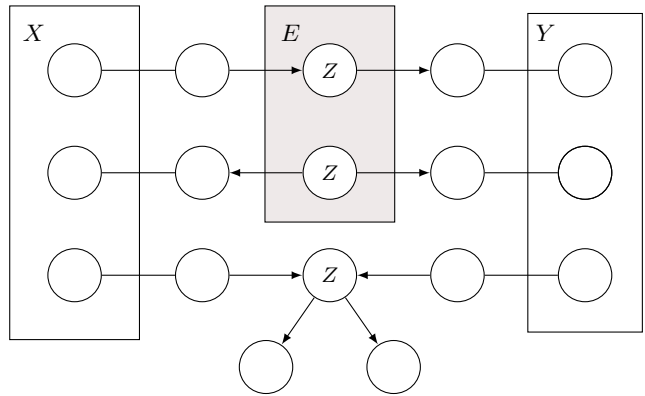

\begin{figure}[H]
	\centering
	\tikzstyle{markov}=[vertex,fill=Snow2]
	\begin{tikzpicture}[every label/.style={font=\footnotesize}]
	\tikzset{
		between/.style args={#1 and #2}{
			at = ($(#1)!0.5!(#2)$)
		}
	}
	\node[vertex] (X) at (0,0) {$X$};
	\node[markov, above left=1.5 of X] (U1) {$U_1$};
	\node (dots1) at (X|-U1) {$\cdots$};
	\node[markov, above right=1.5 of X] (Um) {$U_m$};
	\node[markov, below left=1.5 of X] (Y1) {$Y_1$};
	\node (dots2) at (X|-Y1) {$\cdots$};
	\node[markov, below right=1.5 of X] (Yn) {$Y_n$};
	\node[markov, left=3.5 of X] (Z11) {$Z_{11}$};
	\node[markov, left=1.5 of X] (Z1k) {$Z_{1k}$};
	\node[between = Z11 and Z1k] (dots3) {$\cdots$};
	\node[markov, right=1.5 of X] (Zn1) {$Z_{n1}$};
	\node[markov, right=3.5 of X] (Znl) {$Z_{nl}$};
	\node[between = Zn1 and Znl] (dots4) {$\cdots$};
	\path
	(U1) edge[e] (X)
	(Um) edge[e] (X)
	(X)  edge[e] (Y1)
	(X)  edge[e] (Yn)
	(Z11)  edge[e] (Y1)
	(Z1k)  edge[e] (Y1)
	(Zn1)  edge[e] (Yn)
	(Znl)  edge[e] (Yn)
	;
	\end{tikzpicture}
	\caption{Markov blanket of node $X$, consisting of its parent nodes $\{U_1, \dots, U_n\}$, its child nodes $\{Y_1, \dots Y_n\}$ and the other parent nodes of its child nodes $\{Z_{11},\dots,Z_{1k},Z_{21},\dots,Z_{n1},\dots,Z_{nl}\}$~\cite{russel}.}\label{Markov-Decke}
\end{figure}
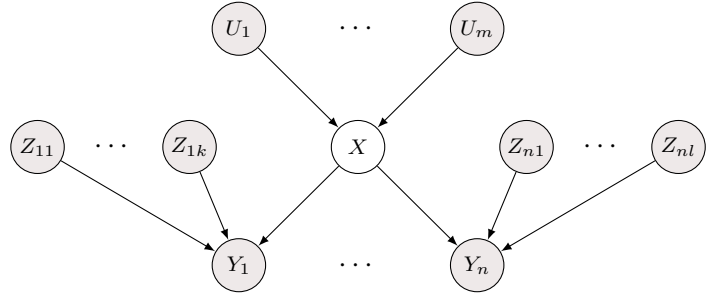

The graph structure of the network encodes statements of conditional independence. Two or more independence statements may imply another independence statement. The rules applicable here are defined as graphoid axioms.\footnote{The predicate \textit{Independence} $I(X,Z,Y)$ is defined as $X$ is independent of $Y$ with observed $Z$ ($Z$ is part of the evidence set~$E$). The graphoid axioms then are:
\begin{enumerate}
		\item Symmetry: $I(X,Z,Y) \Rightarrow I(Y,Z,X)$.
		\item Decomposition: $I(X,Z,Y \cup W) \Rightarrow I(X,Z,Y) \wedge I(X,Z,W)$.
		\item Weak Union: $I(X,Z,Y \cup W)\Rightarrow I(X,Z \cup Y,W)$.
		\item Contraction: $I(X,Z,Y) \wedge I(X,Z \cup Y,W) \Rightarrow I(X,Z,Y \cup W)$.
		\item Intersection: $I(X,Z \cup W, Y) \wedge I(X,Z \cup Y, W) \Rightarrow I(X,Z,Y \cup W)$.
\end{enumerate}
A graphoid is a dependency model $M$ which is completed under rules 1-5, a semi-graphoid is completed under rules 1-4.
	
}\@ D-separability (figure~\ref{fig:d-separation}) is a graph-theoretical relation that covers all these derivable independencies. $Z$ d-separates $X$ and $Y$ in a DAG if within the network $X$ and $Y$ are independent with respect to the graphoid axioms when $Z$ is known. From this the following relations can be derived:

\begin{itemize}
	\item Each node is independent of its non-successors for given parents.
	\item Each node is independent of all other nodes in the network if its Markov blanket (figure~\ref{Markov-Decke}) is known.
\end{itemize}

In figure~\ref{fig:d-separation}, nodes from the set $X$ are independent of nodes from the set $Y$ if one of the three criteria is met for all paths between $X$ and $Y$. For criteria (1) and (2), $Z$ must be known (included in the evidence set $E$). For criterion (3), neither $Z$ nor any of its successors must be included in the evidence set $E$. Non-directional edges are edges for which the direction is irrelevant for the fulfillment of the criterion~\cite{RuNo95}.

As will be explained in section~\ref{chapter:theory:sec:inference}, the independence statements in Bayesian networks are important to reduce the complexity of inference algorithms.

\subsection{Representation of conditional distributions}

The conditional distributions can be denoted as truth tables. However, this becomes inefficient with only a few parameters, since the combination possibilities of states of parent and child nodes and thus the number of rows in a truth table increase exponentially.  More compact options for the representation of local relations between the random variables are needed instead. Canonical probability models are widely in use, which can model larger conditional distributions with few parameters~\cite{Diez00canonicalprobabilistic}. In the following, the relationships that were used to model the prototypical network and can be processed by the implemented agent are explained. An extension of the agent by other canonical models is possible with manageable effort. 

\subsubsection{Complete conditional probability table}

The simplest case is a complete conditional probability table (CPT). For each combination of possible states of a child node and its parent nodes a probability must be found. This combination of states, together with the probability, forms a row in the table. Thus the direct notation of such a table becomes impracticable with only a few parent nodes.   

Canonical probability models are a way of representing a conditional probability distribution with few parameters. Using the respective formula of the canonical distribution, a complete conditional probability table is constructed from these parameters.

\subsubsection{Deterministic node}\label{sec:deterministic_var}

The simplest case of a relation between parent and child nodes is a deterministic relation, i.\,e.\ one that does not contain any uncertainty. The value of the child node is determined by a function that accepts the values of the parent nodes as parameters. 

\begin{Def}[Conditional probability distribution]
	The conditional probability is as follows:
	\[P(y|\mathrm{\mathbf x}) = \begin{cases}
	1 & \mbox{if } y = f(x) \\
	0 & \mbox{otherwise.}
	\end{cases}\]
\end{Def}

A list of common functions is shown in table~\ref{Deterministische Funktionen}.
\begin{table}[h]
	\centering
	\captionsetup{width=1.0\textwidth}
	\begin{tabular}{ccll}
		\toprule
		Function type & Variable type & Name  & Definition \\ \toprule 
		logical      & boolean      & NOT   & $y\Longleftrightarrow \neg x$ \\
		logical      & boolean      & OR    & $y\Longleftrightarrow x_1 \vee \ldots \vee x_n$ \\
		logical      & boolean      & AND   & $y\Longleftrightarrow x_1 \wedge \ldots \wedge x_n$ \\ \midrule
		algebraic  & ordered     & MINUS & $y\Longleftrightarrow -x$ \\
		algebraic  & ordered     & INV   & $y\Longleftrightarrow x_{\text{max}}-x$ \\
		algebraic  & ordered     & MAX   & $y\Longleftrightarrow \text{max}(x_1,\ldots,x_n)$ \\
		algebraic  & ordered     & MIN   & $y\Longleftrightarrow \text{min}(x_1,\ldots,x_n)$ \\
		\bottomrule
	\end{tabular}
	\caption{Definitions of some deterministic functions.}\label{Deterministische Funktionen}
\end{table}

\subsubsection{Stochastical links (ICI models)}\label{sec:ICI models}

In practice, deterministic relationships are not very frequent, since the interaction of different influencing factors in the real world is usually characterized by uncertainty. If, however, for the parent nodes of a relationship in a Bayesian network it is assumed that they are independent in their causal influence on a child node, a number of models for probabilistic relationships which manage with only a few parameters can be formulated. From these parameters and with the corresponding formulas a complete conditional probability table can be constructed. These \textbf{ICI models}\footnote{Diez and Druzdzel speak of \textit{"independence of causal influence"}, hence the expression \textit{"ICI"}~\cite{Diez00canonicalprobabilistic}.}~\cite{Diez00canonicalprobabilistic} can be divided into two classes: Noisy and leaky. Both function according to the same principle, whereby in the case of a noisy model it is assumed that the issue under consideration is completely covered by the model. In the leaky model, however, an additional leak parameter is estimated, which covers the not explicitly modelled cases. In the following, first the two basic ICI classes are explained and then three concrete noisy ICI models are derived. 

\paragraph{Noisy ICI models}\label{sec:ICI models:noisy}

Noisy ICI models can be developed theoretically from deterministic models by introducing auxiliary variables ${Z_1, \dots, Z_n}$, where $n$ is the number of parent nodes. These auxiliary variables serve only to explain the model, but are not part of the modelling. $Y$ is a deterministic function of the $Z_i$s, while each $Z_i$ depends probabilistically on $X_i$, represented by a CPT $P(z_i|x_i)$ each. These CPTs are the parameters of the model. The conditional probability $P(y|\mathrm{\mathbf{x}})$ is calculated as:   
\[P(y|\mathrm{\mathbf{x}}) = \sum_{\mathrm{\mathbf{z}}} P(y|\mathrm{\mathbf{z}}) \cdot P(\mathrm{\mathbf{z}}|\mathrm{\mathbf{x}}),\]
where $P(\mathrm{\mathbf{z}}|\mathrm{\mathbf{x}})$ is the product of all the $P(z_i|x_i)$ :
\[P(\mathrm{\mathbf{z}}|\mathrm{\mathbf{x}}) = \prod_i P(z_i|x_i).\]
For all noisy ICI models it holds true that
\begin{align}\label{eq:noisy_ici}
	P(y|\mathrm{\mathbf{x}}) = \sum_{\mathrm{\mathbf{z}}|f(\mathrm{\mathbf{z}})=y} \prod_i P(z_i|x_i).
\end{align}

\paragraph{Leaky ICI models}\label{sec:ICI models:leaky}
The noisy ICI model assumes that all variables that influence the node $Y$ are explicitly included in the model. However, this modelling is not always feasible or desirable for real-world problems. In a reduced model, only those variables of the system are explicitly contained for which the parameters can be determined well from the available data or which are particularly relevant for the modelled situation.  The influence of implicit variables $\mathrm{\mathbf{ V_I}}$ on $Y$ can be estimated by the leak parameter $P(z_L)$, where $Z_L$ is a virtual variable that summarizes the effect of $\mathrm{\mathbf{V_I}}$.
\begin{align}
	P(y|\mathrm{\mathbf{x}}) = \sum_{\mathrm{\mathbf{z}}|f(\mathrm{\mathbf{z}})=y} \prod_{i|X_i \in \mathrm{\mathbf{X}}} P(z_i|x_i) \sum_{z_L|f(\mathrm{\mathbf{z}},z_L)=y}P(z_L)
\end{align}

\paragraph{Noisy OR}\label{sec:ICI models:noisy:or}
With the noisy OR, $Y$ is an uncertain OR link over its parent node $X_i$. If a noisy OR link is interpreted causally, every $X_i$ is a possible cause of $Y$. The conditional probabilities $Z_i$ indicate whether $Y$ was generated by $X_i$. The term "noisy" thus refers to the possibility that a present cause $x_i$ does not produce an effect $y$. For the model described here, a $\neg z_i$ means that $X_i$ did not lead to $Y$, no matter if it was not present with $X_i=\neg x_i$ or if an inhibitor $I_i$ prevented it from generating $Y=y$. With $q_i$ the probability is indicated that such an inhibitor $I_i$ is active.\footnote{Events become more and more probable if several conditions are fulfilled. For example, if $A$ causes a heart attack with a probability of $c_A=1-q_A$ and $B$ causes a heart attack with $c_B=1-q_B$, then a probability of $1-q_A q_B$ results for a heart attack and for the presence of both diseases. This expression is greater than $c_A$ and $c_B$.} 
\begin{Bsp}[Noisy OR]
	If a heart attack ($I$) is the result of arteriosclerosis (calcification of the blood vessels) ($A$), hypertension (high blood pressure) ($H$) or endocarditis (inflammation of the inner lining of the heart) ($E$), then
	\[q_A = P(\neg i|a,\neg h, \neg e) = 0,1,\]
	means that the occurrence of a heart attack that was caused by vessel calcification was prevented causally independent of the other causes.
	The following applies to the other two causes:
	\begin{align*}
		q_H &= P(\neg i|\neg a, h, \neg e) = 0,2\text{ and} \\
		q_E &= P(\neg i|\neg a, \neg h, e) = 0,6.
	\end{align*}
\end{Bsp}

The probability that $Y$ was generated by $X_i$ is 
\begin{align}\label{eq:noisy_or_+z+x_parameter}
	c_i = P(+z_i|+x_i) = 1 - q_i.
\end{align}
If $X_i$ does not occur, it cannot cause $Y$ either, which is why
\begin{align}\label{eq:noisy_or_+z-x_parameter}
	P(+z_i|\neg x_i) = 0
\end{align}
applies.

From~\ref{eq:noisy_or_+z+x_parameter} and~\ref{eq:noisy_or_+z-x_parameter} the parameter table for noisy OR follows\@:
\begin{table}[H]
	\small
	\centering
	\captionsetup{width=0.8\textwidth}
	\begin{tabular}{|c|cc|}
	\hline	
		$P(z_i|x_i)$ & $+x_i$  & $\neg x_i$ \\ \hline
		$+z_i$       & $c_i$   & $0$ \\ 
		$\neg z_i$   & $1-c_i$ & $1$ \\ \hline
	\end{tabular}
	\caption{Parameters in CPT form for noisy OR and for $X_i\rightarrow Y$}\label{Parameter-CPT Noisy-Or}
\end{table}

To calculate the complete CPT with equation~\ref{eq:noisy_ici}, first define $I_+$ and $I_\neg$ as sets of the indices of the variables which in a configuration of binary parent nodes are \textit{true} or \textit{false}, respectively. Formally, this is expressed as follows:
\begin{align}\label{Bool-Indizes +}
I_+(\prb v)&=\{i|V_i \mbox{ assumes }+v_i\mbox{ in }\prb v\}\\\label{Bool-Indizes -}
I_\neg(\prb v)&=\{i|V_i \mbox{ assumes }\neg v_i\mbox{ in }\prb v\}.
\end{align}

Considering the fact that $f_\text{OR}(\prb z)=\neg y$ is only valid for the configuration ($\neg z_1,\ldots,\neg z_n$), from equation~\ref{eq:noisy_ici} results:
\[P(\neg y|\prb x)=\prod_{i=1}^n P(\neg z_i|x_i)=\prod_{i\in I_+(\prb x)} P(\neg z_i|+x_i)\cdot \prod_{i\in I_\neg(\prb x)} P(\neg z_i|\neg x_i).\]

When inserting the parameters from table~\ref{Parameter-CPT Noisy-Or}, the calculation rule for the complete conditional probability distribution of a noisy OR operation is 
\begin{align}
P(\neg y|\prb x) = \prod_{i\in I_+(x)} q_i = \prod_{i\in I_+(x)}(1-c_i).
\end{align}

The complementary event is used to calculate $P(+y|\prb x)= 1 - P(\neg y|\prb x)$. Table~\ref{tab:noisy-or_coronary_example} shows the complete calculation of a table for the heart attack example.

\begin{table}[H]
	\small
	\centering
	{\renewcommand{\arraystretch}{1.2}
		\begin{tabular}{lllll}
			\toprule
			Endocarditis  	& Hypertension    & Arteriosclerosis  	& $P(Heart Attack)$ & $P(\neg Heart Attack)$    \\
			($E$)			& ($H$)			& ($A$)				& ($P(I)$) 		   & ($P(\neg I)$)  			\\
			\toprule
			\textit{false}  & \textit{false} & \textit{false} & \textbf{0,0} & \textbf{1,0}                  \\ \midrule 
			\textit{false}  & \textit{false} & \textit{true}  & $\mathbf{0,3} = 1 - 0,7$ & \textbf{0,7}                  \\ \midrule 
			\textit{false}  & \textit{true}  & \textit{false} & $\mathbf{0,4} = 1 - 0,6$ & \textbf{0,6}                  \\ \midrule 
			\textit{false}  & \textit{true}  & \textit{true}  & $0,58 = 1 - 0,42$  & $0,42 = 0,6 \cdot 0,7$     \\ \midrule  
			\textit{true}   & \textit{false} & \textit{false} & $\mathbf{0,6} = 1 - 0,4$  & \textbf{0,4}                  \\ \midrule 
			\textit{true}   & \textit{false} & \textit{true}  & $0,72 = 1 - 0,28$ & $0,28 = 0,4 \cdot 0,7$     \\ \midrule  
			\textit{true}   & \textit{true}  & \textit{false} & $0,76 = 1 - 0,24$ & $0,24 = 0,4 \cdot 0,6$     \\ \midrule  
			\textit{true}   & \textit{true}  & \textit{true}  & $0,832 = 1 - 0,168$ & $0,168 = 0,4 \cdot 0,6 \cdot 0,7$ \\ \bottomrule  
	\end{tabular}}
	\caption{Causes of heart attack as an example for noisy OR.}\label{tab:noisy-or_coronary_example}
\end{table}

A heart attack ($I$) can be caused causally independent of endocarditis ($E$), hypertension ($H$) and arteriosclerosis ($A$). The probabilities that a heart attack will not occur under one of these causes are $q_E = 0.4$, $q_H = 0.6$ and $q_A = 0.7$. These inhibitor probabilities can be used to calculate the complete CPT for $P(I|A,H,E)$. The values that occur in a parameter-CPT of a noisy OR are marked bold in table~\ref{tab:noisy-or_coronary_example}.

\paragraph{Noisy MAX}\label{sec:ICI models:noisy:max}
The noisy MAX model represents a generalization of the noisy OR for a variable $Y$ with more than two states. Here $Z_i$ represents the value of $Y$ caused by $X_i$. The largest $z_i$ yields the resulting value of $Y$, so $y=f_\text{MAX}(\prb z)$ applies. All $Z_i$s must have the same domain as $Y$. The noisy MAX parameters for the connection $X_i\rightarrow Y$ are
\begin{align}
	c_{z_i}^{x_i} = P(z_i|x_i),
\end{align}
where each $c_y^{x_i}$ can be understood as a probability that $X_i$ with the value $x_i$ sets the value of $Y$ to $y$. The complete CPT for a noisy MAX model is determined by first calculating $P(Y\leq y|\prb x)$ for all values of $y$ and all configurations $\prb x$:
\[P(y\leq y|\prb x)=\sum_{\prb z|f_{\mathrm{MAX}}(\prb z)\leq y}\prod_i c_{z_i}^{x_i}=\sum_{z_1 \leq y} \cdots \sum_{z_n\leq y}\prod_i c_{z_i}^{x_i}=\prod_i\left(\sum_{z_i\leq y} c_{z_i}^{x_i}\right).\]
This results in the following calculation for the accumulated parameters:
\begin{align}\label{eq:noisy_max:accumulated_parameter}
	C_y^{x_i}=\sum_{z_i\leq y} c_{z_i}^{x_i}
\end{align}
and therefore
\begin{align}\label{eq:noisy_max:accu_parameter_product}
	P(Y\leq y|\prb x)=\prod_i C_y^{x_i}.
\end{align}
The CPT values can now be calculated as follows:
\begin{align}\label{eq:noisy_max:cpt_calculation}
	P(y|\prb x) = \begin{dcases}
		P(Y\leq y|\prb x) - P(Y\leq y-1|\prb x), & \mbox{ if }y \neq y_{\min} \\
		P(Y\leq y|\prb x), & \mbox{ if }y = y_{\min}
	\end{dcases}
\end{align}

\paragraph{Noisy AND}\label{sec:ICI models:noisy:and}
The parent nodes of a noisy AND link can be interpreted as conditions that must be fulfilled for $Y$ to become true, where the uncertainty is created by the fact that any condition can be suppressed or replaced. Inhibitor probabilities work as in the case of noisy OR\@: $q_i$ is the probability that $I_i$ is active, and again $c_i = 1 - q_i$ applies. If no inhibitor is active, then $c_i=1$. The probability that the $i$th substitute replaces $X_i$ if the condition is not met is called $s_i$, and $s_i=0$ if there is no substitute for $X_i$. In general, $c_i \cong 1$ and $s_i \cong 0$ apply. The following CPT of a noisy AND parameter $P(z_i|x_i)$ results:
\begin{table}[H]
	\small
	\centering
	\captionsetup{width=0.8\textwidth}
	\begin{tabular}{|c|cc|}
	\hline	
		$P(z_i|x_i)$ & $+x_i$  & $\neg x_i$ \\ \hline
		$+z_i$       & $c_i$   & $s_i$ \\ 
		$\neg z_i$   & $1-c_i$ & $1-s_i$ \\ \hline
	\end{tabular}
	\caption{Parameters in CPT form for noisy AND and the connection \mbox{$X_i\rightarrow Y$}.}\label{Parameter-CPT Noisy-AND}
\end{table}
The total CPT is calculated according to the following formula:
\[P(+y|\prb x) = \prod_i P(+z_i|x_i) = \prod_{i\in I_+(\prb x)}c_i\prod_{j\in I_\neg(\prb x)}s_j.\]

\section{Inference Mechanisms}\label{chapter:theory:sec:inference}

Inference is the most important operation in Bayesian networks. It is dedicated to the question which probability distribution a variable $X$ assumes under observation of certain values for other variables. The evidence $e$ is given as a set of variables instantiated with observed values. The variable $X$ represents the query variable. The posteriori probability with $P(X=x|e)$ is searched for\footnote{The calculation of the posteriori probabilities is very difficult in terms of complexity theory. The efficiency of the algorithms depends on the properties of the network. With optimal network modelling, however, for real-world problems an efficient calculation is possible.}.

First, with the method of summation a simple and exact inference method is presented for illustration purposes (section~\ref{sec:sum_up_inference}). However, it will be shown that the calculation effort with this method increases exponentially with the number of variables. More efficient methods are therefore interesting for practical usage, which will be presented in the following:
\begin{itemize}
\item The algorithm of belief propagation (section~\ref{sec:BP}) is efficient and exact for simply connected networks and there are approaches with the inference in cluster trees (section~\ref{sec:Junction-Tree-Inference}) and loopy belief propagation (section~\ref{sec:Loopy-Belief-Propagation}) which make it usable also for multiply connected networks~\cite{Pearl82,Pearl83}. 
\item Gibbs sampling is an application of Monte Carlo Markov chain methods (MCMC) to the inference in Bayesian networks (section~\ref{sec:approx_inference})~\cite{Geman84}. 
\item Variational message passing is a modern algorithm for efficiently approximating even complex distributions (section~\ref{chapter:theory:sec:inference:subsec:VMP})~\cite{JGJS99}. 
\end{itemize}

\subsection{Inference by summation}\label{sec:sum_up_inference}

Each conditional probability can be calculated by adding up the terms from the joint distribution. For a single query variable $X$, this is done using the equation
\[\label{cond_prob_as_sum}
\prb{P}(X|\prb{e}) = \alpha\prb{P}(X,\prb{e}) = \alpha\sum_y \prb{P}(X,\prb{e},\prb{y}),\]
where $\alpha$ is a normalization factor and $\prb y$ are instances of all unobserved variables $\prb{Y}$ (without $X$).

\begin{Alg}[Inference by summation]
	The algorithm runs in two steps:
	\begin{enumerate}
		\item Using the equation of the joint density and applying the product rule, the terms $P(x,\prb{e},\prb{y})$ are calculated as products from the conditional probabilities of the network for each possible assignment of the unobserved variables.
		\item Subsequently the individual probabilities are summed up.
	\end{enumerate}
\end{Alg}

The algorithm for inference by summation can be represented in pseudo code as follows~\cite{russel}.

\begin{Alg}[Pseudocode for inference by summation]\leavevmode
	\begin{algorithmic}[1]
		\Function{Enumeration-Ask} {$X,\textbf{e},bn$}
			\State{\Return{a distribution over $X$}}
		  \State{Input $X$, the query variable}
		  \State{Input $\textbf{e}$, observed values for the variables $\textbf{E}$}
		  \State{Input $\textbf{Y}$, all unobserved variables}
		  \State{Input $bn$, a Bayesian network with variables $\{X\}\cup{}\textbf{E}\cup{}\textbf{Y}$}
		  \State{$P(X|\textbf{e}) \gets$ a distribution over X, initially empty}
		  \ForAll{Value $x$ of $X$}
		    \State{Expand $\textbf{e}$ with value $x$ for $X$}
				\State{$P(x) \gets{}\Call{Enumeration-All}{\Call{Vars}{bn}, \textbf{e}}$}
		  \EndFor{}
			\State{\Return{\Call{Normalize}{$P(X|\textbf{e})$}}}
		\EndFunction{}
	\end{algorithmic}
	
	\begin{algorithmic}[1]
		\Function{Enumeration-All} {$vars,\textbf{e}$} \State{\Return{a real number}}
		\If{\Call{IsEmpty}{$vars$}}
			\State{\Return{$1.0$}}
		\EndIf{}
		\State{$Y \gets{}\Call{First}{vars}$}
		\If{$Y$ has the value $y$ in $\textbf{e}$}
			\State{\Return{$P(y | parents(Y)\times{}\Call{Enumeration-All}{\Call{Rest}{vars},\textbf{e}}$}} 
		\Else{}
			\State{\Return{$\sum_{y} P(y | parents(Y) \times{}\Call{Enumeration-All}{\Call{Rest}{vars},\textbf{e}_{y}}$}}
			\State{where $\textbf{e}_{y}$ equals $\textbf{e}$ expanded with $Y=y$}
		\EndIf{}
		\EndFunction{}
	\end{algorithmic}
\end{Alg}

This method is equivalent to the depth search in a tree and thus has a linear storage complexity, but an exponential time complexity.

If the calculation is visualized in an expression tree, it becomes clear that many partial calculations are carried out several times. Saving the results of these calculations temporarily for later reuse significantly reduces the required computing time. Likewise, all variables that are not ancestors of a query or evidence variable are irrelevant to the query and can be removed before the calculation. An algorithm that does this is the so-called variable, which is  described in~\cite{russel}.

\subsection{Inference by belief propagation}\label{sec:BP}

Belief propagation is an inference method that makes it possible to efficiently calculate posteriori probabilities for all nodes of the network. However, the basic algorithm is limited in that it can only be applied to simply connected, i.e. polytree-shaped, networks. First, the core idea of the algorithm, the so-called message passing, will be shown. Then a description of how this principle can be transferred to more complex networks follows. Finally, approaches are discussed with which the limitations of the basic algorithm are partially overcome.

\subsubsection{Message passing}\label{chapter:theory:sec:inference:subsec:BP:subsubsec:message_passing_on_chains}
The basic idea of belief propagation is to iteratively update the conditional probability distribution of a node based on its neighbouring nodes. This principle will first be demonstrated on a very simple network. The network has a chain-like structure, the nodes $X_1, X_2,\dots, X_{n-1}, X_n$ are lined up linearly.
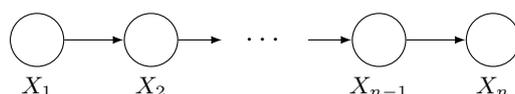
\begin{figure}[H]
	\centering
	\begin{tikzpicture}[every label/.style={font=\footnotesize}]
	\node[vertex] (1) at (0,0) [label=below:$X_1$]{};
	\node[vertex] (2) at (1.5,0) [label=below:$X_2$]{};
	\node (dots) at (3,0) {$\cdots$};
	\node[vertex] (3) at (4.5,0) [label=below:$X_{n-1}$]{};
	\node[vertex] (4) at (6,0) [label=below:$X_n$]{};
	\path
	(1) edge[e] (2)
	(2) edge[e,shorten >=5pt] (dots)
	(dots) edge[e,shorten <=5pt] (3)
	(3) edge[e] (4);
	\end{tikzpicture}
\end{figure}

\textbf{Factorisation of the joint density distribution} \\
The marginal distribution\footnote{The marginal distribution is formed by the marginal probabilities. The name originates from the marginal frequency, which is the sum of the frequencies of a characteristic at the edge of a contingency table. Marginal probabilities are calculated according to the same principle by summing the conditional probabilities of a variable. This process is also called marginalization.} of any variable $X_i$ can be calculated by adding up all possible values of all other variables:
\[P(X_i)=\sum_{X_1}\sum_{X_2}\cdots\sum_{X_{i-1}}\sum_{X_{i+1}}\cdots\sum_{X_n}P(\prb X).\]

\textbf{Definition of the messages} \\
It is now possible to calculate $\mu_\beta$ as the message to be transmitted between the nodes iteratively on the node chain back to the desired node $X_i$:
\begin{align*}
	\mu_\beta(X_{n-2}) &= \sum_{X_{n-1}} P(X_{n-1}|X_{n-2})\mu_\beta(X_{n-1}) \\
	& \cdots \\
	\mu_\beta(X_i) &= \sum_{X_{i+1}} P(X_{i+1}|X_i)\mu_\beta(X_{i+1}).
\end{align*}
The same can be done beginning from the other end of the chain:
\begin{align*}
	\mu_\alpha(X_2) &= \sum_{X_1}P(X_1)P(X_2|X_1) \\
	& \cdots \\
	\mu_\alpha(X_i) &= \sum_{X_{i-1}}P(X_i|X_{i-1})\mu_\alpha(X_{i-1}).
\end{align*}
The marginal distribution $P(X_i)$ can now be calculated as the product of the contributions coming from both ends:
\[P(X_i)=\mu_\alpha(X_i)\mu_\beta(X_i).\]

Thus $\mu_\alpha(X_i)$ can be interpreted as a message transmitted from node $X_{i-1}$ to node $X_i$, and $\mu_\beta(X_i)$ as a message from node $X_{i+1}$ to node $X_i$.
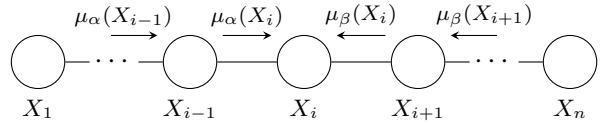
\begin{figure}[H]
	\centering
	\begin{tikzpicture}[every label/.style={font=\footnotesize}]
	\node[vertex] (1) at (0,0) [label=below:$X_1$]{};
	\node (dots1) at (1,0) {$\cdots$};
	\node[vertex] (2) at (2,0) [label=below:$X_{i-1}$]{};
	\node[vertex] (3) at (3.5,0) [label=below:$X_{i}$]{};
	\node[vertex] (4) at (5,0) [label=below:$X_{i+1}$]{};
	\node (dots2) at (6,0) {$\cdots$};
	\node[vertex] (5) at (7,0) [label=below:$X_n$]{};
	\path
	(1) edge[shorten >=-2pt] (dots1)
	(dots1) edge[shorten <=-2pt] (2)
	(2) edge (3)
	(3) edge (4)
	(4) edge[shorten >=-2pt] (dots2)
	(dots2) edge[shorten <=-2pt] (5);
	\path
	(dots1.west) edge[message arrow,shorten <=10pt] node[message]{$\mu_\alpha(X_{i-1})$} (2.west)
	(2.east)     edge[message arrow]                node[message]{$\mu_\alpha(X_{i})$} (3.west)
	(3.east)     edge[message arrow,<-]             node[message]{$\mu_\beta(X_{i})$} (4.west)
	(4.east) edge[message arrow,<-,shorten >=10pt]  node[message]{$\mu_\beta(X_{i+1})$} (dots2.east);
	\end{tikzpicture}
\end{figure}

Each outgoing message is obtained by multiplying the incoming message by the local conditional probability and summing up all possible values of the sending node. \\

\textbf{Complete message passing} \\
To calculate the marginal probabilities of all nodes $X_i$ of the chain, the following scheme is used:
\begin{enumerate}
	\item Sending the messages $\mu_\alpha$, starting with $\mu_\alpha(X_1)$ to $\mu_\alpha(X_n)$
	\item Sending back messages $\mu_\beta$, starting with $\mu_\beta(X_n)$ and ending with $\mu_\beta(X_1)$
\end{enumerate}

\textbf{Considering evidence} \\
If the posteriori probability is to be calculated under given observations $\prb e$, only the observed value is used when calculating the messages of the observed node, instead of summing up all possible values of the node. Subsequently, a normalization must be performed:
\begin{align}
	\prb P(X_i|\prb e)=\frac{\prb P(X_i,\prb e)}{\sum_{X_i}P(X_i,\prb e)}.
\end{align}

In section~\ref{sec:message_passing_on_factor_graphs} the message passing scheme described above is generalized to any polytree network. However, this requires a more suitable graph structure, which will be introduced in the next section.

\subsubsection{Factor graphs}

As already shown in the last section, belief propagation uses the factorization of Bayesian networks for an iterative calculation. Therefore it can be implemented and explained more easily on a graph structure that emphasizes this factorization.  

\begin{figure}[H]
	\centering
	\scalebox{0.7}{\begin{tikzpicture}[every label/.style={font=\small}]
		\tikzstyle{vertex}=[circle, draw, font=\small, inner sep=0pt, minimum size=25pt]
		\tikzstyle{phantom}=[vertex,draw=none]
		\tikzstyle{factor}=[draw, fill, inner sep=0pt, minimum size=15pt]
		\tikzstyle{description}=[decorate,below, inner sep= 30pt, align=left]
		\node (A) at (0,0){
			\begin{tikzpicture}
			\node[vertex] (g1_x1) at (0,2.85) [label=above left : $X_1$]{};
			\node[vertex] (g1_x2) at (3,2.85) [label=above right: $X_2$]{};
			\node[vertex] (g1_x3) at (1.5,0) [label=below right: $X_3$]{};
			\path
			(g1_x1) edge[e] (g1_x3)
			(g1_x2) edge[e] (g1_x3);
			\draw[description] (g1_x3) node{$P(X_3|X_2,X_1)P(X_1)P(X_2)$};
			\end{tikzpicture}};
		\draw[decorate,font=\large] (A.north west) node[above right=-4pt]{a)};
		
		\node[below right=0 and -1.5 of A.north east] (B){
			\begin{tikzpicture}
			\node[vertex] (g2_x1) at (0,3)   [label=above left : $X_1$]{};
			\node[vertex] (g2_x2) at (3,3)   [label=above right: $X_2$]{};
			\node[phantom] (g2_v1) at (1.5,3) {}; 
			\node[factor, below=0.5 of g2_v1] (g2_f)  [label=above      : $f$]{};
			\node[vertex, below=0.9 of g2_f]  (g2_x3) [label=below right: $X_3$]{};
			\path
			(g2_x1) edge (g2_f)
			(g2_x2) edge (g2_f)
			(g2_x3) edge (g2_f);
			\draw[description] (g2_x3) node{$f(X_1,X_2,X_3)=P(X_3|X_2,X_1)P(X_1)P(X_2)$};
			\end{tikzpicture}};
		\draw[decorate,font=\large] (B.north west) node[above right=-4pt and 30pt]{b)};
		
		\node[below right=0 and -1.5 of B.north east] (C){
			\begin{tikzpicture}
			\node[vertex] (g3_x1) at (0,3)            [label=above left : $X_1$]{};
			\node[vertex] (g3_x2) at (3,3)            [label=above right: $X_2$]{};
			\node[phantom] (g3_v1) at (1.5,3) {}; 
			\node[factor, below=0.5 of g3_x1] (g3_fa) [label=left       : $f_a$]{};
			\node[factor, below=0.5 of g3_x2] (g3_fb) [label=right      : $f_b$]{};
			\node[factor, below=0.5 of g3_v1] (g3_fc) [label=above      : $f_c$]{};
			\node[vertex,below=0.9 of g3_fc]  (g3_x3) [label=below right: $X_3$]{};
			\path
			(g3_x1) edge (g3_fa)
			(g3_x2) edge (g3_fb)
			(g3_x1) edge (g3_fc)
			(g3_x2) edge (g3_fc)
			(g3_x3) edge (g3_fc);
			\draw[description] (g3_x3) node{$f_a(X_1)=P(X_1)$\\
				$f_b(X_2)=P(X_2)$\\
				$f_c(X_1,X_2,X_3)=P(X_3|X_2,X_1)$};
			\end{tikzpicture}};
		\draw[decorate,font=\large] (C.north west) node[above right=-4pt and 10pt]{c)};
	\end{tikzpicture}}
	\caption{Bayesian network and equivalent factor graphs.}\label{fig:factor_graph}
\end{figure}
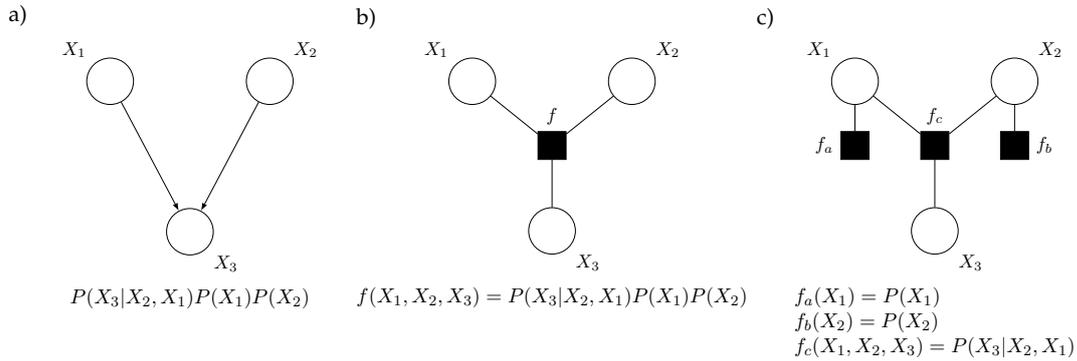

\begin{Def}[Factor graph]
A factor graph is a bipartite graph\footnote{Bipartite graphs describe the relationships between the elements of two sets.} with the following properties:
	\begin{itemize}
		\item It consists of two different node types, variables and factors, and undirected edges between a variable and a factor.\footnote{As with the terms \textit{node} and \textit{variable} in Bayesian networks, in factor graphs the terms \textit{factor} and \textit{factor node} are used synonymously, whereby the first term is used more commonly in a statistical context and the second one in the context of graph structures. The term \textit{variable} or \textit{variable node} is used for the nodes of the variable set.}
		\item A node of the variable set represents a random variable $X_i$ and is represented by a circle.
		\item A node of the factor set represents a factor $f_j$, represented by a square. It describes a probability distribution over the adjacent random variables. This can be a conditional probability distribution, but also a product of probability distributions or even a joint probability distribution (figure~\ref{fig:factor_graph} b) and c)).		
	\end{itemize}
\end{Def}

The Bayesian network in figure~\ref{fig:factor_graph} a) can be converted into equivalent factor graphs. The a-priori probabilities of the nodes $X_1$ and $X_2$ can b) be summarized directly in a single factor or c) modelled in separate factors.

Every Bayesian network can be represented by a factor graph.\footnote{The corresponding proof is based on the so-called Hammersley-Clifford theorem~\cite{DBLP:journals/corr/abs-1207-1366}.} The procedure for constructing such an equivalent factor graph is as follows:
\begin{enumerate}
	\item A variable node is created for each node of the Bayesian network.
	\item  For each child node of the Bayesian network, a factor node with the corresponding conditional probability distribution is generated. The factor node is then connected by undirected edges to the equivalent variable nodes of the child node's parent node and the child node itself.	
	\item For each root node, a factor is generated with the a-priori probability of the respective root node and connects it with the equivalent variable of the root node by an undirected edge (figure~\ref{fig:factor_graph} c).\\
	Alternatively, the a-priori probability is added to each factor node adjacent to the variable node which is equivalent to the root node by multiplying the probability by this factor (figure~\ref{fig:factor_graph} b).
\end{enumerate}

\subsubsection{Sum-product algorithm on factor graphs}\label{sec:message_passing_on_factor_graphs}

The sum-product algorithm is an efficient method for exact inference on graphs with a tree-like structure. It is a generalization of the message-passing algorithm previously shown for the chain-like network.

\begin{figure}[H]
	\centering
	\scalebox{0.8}{
		\begin{tikzpicture}[every label/.style={font=\small}]
		\tikzstyle{vertex}=[circle, draw, font=\scriptsize, inner sep=0pt, minimum size=25pt]
		\tikzstyle{factor}=[draw, fill, inner sep=0pt, minimum size=15pt]
		\tikzstyle{description}=[decorate,below, inner sep= 30pt, align=left]
		\tikzstyle{phantom}=[vertex,draw=none]
		\tikzstyle{dots}=[loosely dotted,shorten <=3pt, shorten >=3pt]
		\tikzstyle{message}=[above,font=\footnotesize,sloped, inner sep=13pt]
		\tikzstyle{message arrow}=[->,>=stealth,
		shorten >=10pt, shorten <=10pt, 
		decoration={sl,raise=8pt},decorate]
		
		\node[vertex] (x1) at (0,0){};
		\node[factor,draw=none,fill=none,below=0.1 of x1] (pf1) {};
		\node[vertex,  below=0.7 of x1] (x2){};
		\node[phantom, below=1.5 of x1] (px) {};
		\node[factor,  right=1.2 of pf1] (f1)[label={above,align=center:$\hspace{70pt}\mathcal N(X)=\{f_1,\dots,f_n\}$\\$f_1$}]{};
		\node[vertex,  right=4 of px] (x) [label={below:$X$}]{};
		\node[vertex,  below=2 of x1] (x3){};
		\node[phantom, below=2.7 of x1] (pf2) {};
		\node[factor,  below=2.3 of f1] (f2)[label={below:$f_n$}]{};
		\node[vertex,  below=3.7 of x1] (x4){};
		
		\path
		(x1) edge[dots] (x2)
		(x2) edge[dots] (x3)
		(x3) edge[dots] (x4)
		(f1) edge[dots] (f2)
		(x1) edge (f1)
		(x2) edge (f1)
		(x3) edge (f2)
		(x4) edge (f2)
		(f1) edge (x)
		(f2) edge (x)
		(f1) edge[message arrow] node[message]{$\mu_{f_1\rightarrow X}(X)$} (x)
		(f2) edge[message arrow] node[message]{$\mu_{f_n\rightarrow X}(X)$} (x);
		\end{tikzpicture}}
	\caption{The marginal distribution $P(X)$ of a node $X$ is calculated by multiplying the incoming node messages $\mu_{f_1\rightarrow X}(X),\ldots \mu_{f_1\rightarrow X}(X)$ of all neighbouring factors $f_1,\ldots,f_n$.}\label{General Message Passing}
\end{figure}
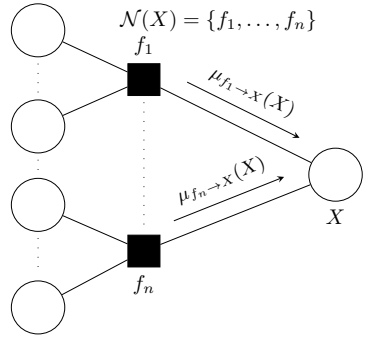

The aim is now to determine the marginal distribution of $X$:
\[\prb P(X)=\sum_{\prb X \backslash X}\prb P(\prb X).\]
If the message passing scheme  for chain-like graphs is generalized, the marginal probability is calculated as the product of the messages coming from all neighbouring factors $f_i \in \mathcal N(X)$ (figure~\ref{General Message Passing}):
\[P(X)=\prod_{f_i\in \mathcal N(X)}\mu_{f_i\rightarrow x}(X).\]

In a factor graph two types of messages are distinguished: the factor message from a factor $f_s$ to a node $X$ and the node message from a node $X_j$ to a factor $f_s$. \\

\textbf{Factor message} \\
A factor message to node $X$ is the product of the incoming messages in the factor node $f_s$ from all neighbouring variable nodes except $X$, multiplied by the factor summed over all possible values of all variables occurring in the factor except $X$:
\begin{align}\label{eq:factor_message}
	\mu_{f_s\rightarrow X}(X)=
	\sum_{X_1}\cdots\sum_{X_m}f_s(X,X_1,\dots,X_m)
	\prod_{X_j\in \mathcal N(f_s)\backslash X} \mu_{X_j\rightarrow f_s}(X_j).
\end{align}

\clearpage

\textbf{Node message} \\
A node message to a factor $f_s$ is the product of the incoming factor messages in the node $X$ from all other factors except $f_s$:
\begin{align}\label{eq:variable_message}
	\mu_{X\rightarrow f_s}(X)=\prod_{f_i\in\mathcal N(X)\backslash f_s} \mu_{f_i\rightarrow X}(X).
\end{align}

\textbf{Initialization} \\
Before the actual message passing algorithm starts, the messages from the leaves of the graph must be initialized. A leaf can be both a factor and a variable. \\

\textbf{Complete message passing} \\
To calculate the marginal probabilities for all nodes of the graph, messages must be sent over all edges of the graph in both of their directions. Similar to message passing on a node chain (section~\ref{chapter:theory:sec:inference:subsec:BP:subsubsec:message_passing_on_chains}) the algorithm runs in two phases. After initializing the leaves, any node of the graph is selected as the root node. 

\begin{enumerate}
	\item In phase 1, starting from the leaf nodes, the messages are sent in the direction of the root node.
	\item In phase 2, starting from the root node, the messages are sent back to the leaf nodes.
\end{enumerate}

If the messages are stored temporarily for each variable node, the marginal probability for each variable can then be determined by multiplying these messages. For the calculation of all marginal probabilities, here again only twice as many messages are required as for the calculation of the marginal probability for a single variable. \\

\textbf{Considering evidence} \\
If the posteriori probabilities are to be calculated under given observations, the observed values of the evidence variables\footnote{Evidence variables are variable nodes of the graph for which a certain value is observed.} are used to calculate factor messages instead of summing up all their values (analogous to section~\ref{chapter:theory:sec:inference:subsec:BP:subsubsec:message_passing_on_chains}). Equation~\ref{eq:factor_message} changes under observation of the node $X_i$ with the value $e_i$ as follows: 
\begin{align}
	\mu_{f_s\rightarrow X}(X)=
	\sum_{X_1}\cdots\sum_{X_{i-1}}\sum_{X_{i+1}}\cdots\sum_{X_m}f_s(X,X_1,\dots,e_i,\dots,X_m)
	\prod_{X_j\in \mathcal N(f_s)\backslash X} \mu_{X_j\rightarrow f_s}(X_j).
\end{align}

The calculation of the posteriori probabilities for the individual nodes $X_i$ under given evidence $\prb e$ is done, just as the calculation of the marginal probabilities without evidence, by multiplying the incoming factor messages, where a normalization with the factor $\frac{1}{\sum_{X}P(X,\prb e)}$ must be carried out.
\begin{align}
	\prb P(X|\prb e) = \frac{\prod_i \mu_{f_i\rightarrow X}(X)}{\sum_X \prod_i \mu_{f_i\rightarrow X}(X)}.
\end{align}

\subsubsection{Exact inference on connected networks with junction trees}\label{sec:Junction-Tree-Inference}

A connected Bayesian network provides a factor graph with loops. Exact inference is no longer possible with the sum-product-algorithm on such a graph (section~\ref{sec:Loopy-Belief-Propagation}). Nonetheless, it is possible to cluster the original Bayesian network into a graph structure where belief propagation with exact results is possible. The transformation algorithm combines nodes that are part of a loop into clusters. The result is a tree-like cluster graph, the so-called junction tree, on which the message passing is then performed.

The junction tree algorithm provides very good results for smaller networks, but its complexity increases exponentially with the maximum number of variables in a cluster, making it unsuitable for large complex networks. Nevertheless, it is a fitting approach for the problem discussed in this report, which has already been shown in~\cite{beleg}. Since it is not implemented in Infer.NET, the practical part however uses a variant of belief propagation described in the following section.

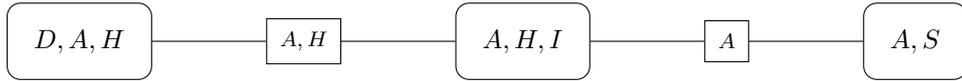
\begin{figure}[H]
	\centering
	\begin{tikzpicture}[every label/.style={font=\small}]
	\tikzstyle{cluster}=[draw, font=\small, rounded corners=5pt, inner sep=10pt]
	\tikzstyle{sepset}=[draw, font=\scriptsize, inner sep=5pt]
	\tikzstyle{description}=[decorate,below, inner sep= 30pt, align=left]
	\tikzstyle{phantom}=[vertex,draw=none]
	\tikzstyle{dots}=[loosely dotted,shorten <=3pt, shorten >=3pt]
	\tikzstyle{message}=[above, inner sep=10pt,font=\footnotesize]
	\tikzstyle{message arrow}=[->,>=stealth,
	shorten >=10pt, shorten <=10pt,
	decoration={sl,raise=8pt},decorate,sloped]
	
	\node[cluster] at (0,0) (DAH)          {$D,A,H$};
	\node[sepset,  right=1.5 of DAH] (AH)    {$A,H$};
	\node[cluster, right=1.5 of AH] (AHI) {$A,H,I$};
	\node[sepset,  right=1.5 of AHI] (A)   {$A$};
	\node[cluster, right=1.5 of A] (AS)  {$A,S$};
	\path
	(DAH) edge (AH)
	(AH) edge (AHI)
	(AHI) edge (A)
	(A) edge (AS)
	;
	\end{tikzpicture}
	\caption{Cluster graph for the network depicted in figure~\ref{Karzinomnetz}.}\label{fig:cluster_graph}
\end{figure}

\subsubsection{Approximate inference with loopy belief propagation}\label{sec:Loopy-Belief-Propagation}

For factor graphs with loops the so-called loopy belief propagation (LBP)~\cite{FMK98,Minka:2001:EPA:647235.720257} algorithm can be applied. It is an iterative procedure, where the belief propagation is repeated until a convergence criterion is reached (algorithm~\ref{alg:loopy_belief_propagation}). Consequently, the results are no longer exact, but an approximation.

\begin{Alg}[Loopy belief propagation] 
Iterative procedure of the loopy belief propagation algorithm:	
	\begin{algorithmic}[1]
		\Function{LBP}{}
		\ForAll{Edges between factors $f$ and variables $X$}
		\State{Initialize $\mu_{f\rightarrow X}^{(0)}$ and $\mu_{X\rightarrow f}^{(0)}$ with uniform distribution}
		\EndFor{}
		\While{$\mu^{(t+1)} \neq \mu^{(t)}$} \Comment{Must be checked for every message $\mu$}
		\ForAll{Edges between factors $f$ and variables $X$}
			\State{$\mu_{f\rightarrow X}^{(t+1)}\leftarrow$\Call{CalcFactorMessage}{$\mu_{X_j \in \mathcal N(f)\backslash X\rightarrow f}^{(t)}$}\Comment{Equation~\ref{eq:LMP_update_1}}}
			\State{$\mu_{X\rightarrow f}^{(t+1)}\leftarrow$\Call{CalcNodeMessage}{$\mu_{f_i\in\mathcal N(X)\backslash f\rightarrow X}^{(t)}$}\Comment{Equation~\ref{eq:LMP_update_2}}}
			\State{$t \gets t+1$}
		\EndFor{}
		\EndWhile{}
		\EndFunction{}
	\end{algorithmic}\label{alg:loopy_belief_propagation}
\end{Alg}

In principle, here an iteration step is the same as for the complete message passing scheme from section~\ref{sec:message_passing_on_factor_graphs}: A message is sent through each edge of the graph in both directions. The difference is that with loopy belief propagation a message is determined by using the messages from the previous iteration step, whereas with ordinary belief propagation the calculation of a message is based on messages from the same (and only existing) iteration step. If $t$ and $t+1$ are the previous and current iteration steps, respectively, the messages in step $t+1$ are calculated as follows:

\begin{align}\label{eq:LMP_update_1}
	\mu_{f\rightarrow X}^{(t+1)}(X)&= \sum_{X_1}\cdots\sum_{X_m}f_s(X,X_1,\dots,X_m) \prod_{X_j\in \mathcal N(f)\backslash X} \mu_{X_j\rightarrow f}^{(t)}(X_j)\\
	\mu_{X\rightarrow f}^{(t+1)}(X)&=\prod_{f_i\in\mathcal N(X)\backslash f} \mu_{f_i\rightarrow X}^{(t)}(X)\label{eq:LMP_update_2}
\end{align}

where $\mu_{f\rightarrow X}^{(t+1)}(X)$ and $\mu_{X\rightarrow f}^{(t+1)}(X)$ are the factor and node messages.

In order to calculate the messages for step $t=1$, they must be initialized for all edges for step $t=0$. Since the messages themselves are probability distributions, they are typically initialized with a uniform distribution. Then with each iteration step all messages are updated with equations~\ref{eq:LMP_update_1} or ~\ref{eq:LMP_update_2}. The iteration continues until the messages converge, i.\,e.\ until 
\[\mu^{(t+1)}=\mu^{(t)}\]
holds true for all messages $\mu$ or another termination criterion (e.g. a maximum number of iterations) is reached. After all, there is no guarantee that the messages will converge. In some cases it is possible that the messages start to oscillate after a certain iteration. What is more, it has been shown that the order in which messages are recalculated in an iteration step affects the convergence. Updating the messages simultaneously leads to oscillation much more often than calculating one after the other. A clever planning of the order in which the messages are calculated also improves the results in terms of their accuracy. Several procedures for this scheduling of the update order exist, some of which are described in~\cite{Sutton_McCallum_Dynamic_Schedules_BP}. An analysis of the conditions under which LBP converges can be found in~\cite{Mooij_Kappen_Sufficient_Condidtions_Convergence_SPA}.

Although the algorithm does not ensure convergence, it has proven itself in many applications\footnote{One of the most popular applications of loopy belief propagation are the turbo codes, which for the first time enabled information coding near the Shannon boundary. They were also the proof of the practical applicability of the procedure~\cite{Turbo-Codes}.} and provides very good results for the type of Bayesian networks described in this thesis. LBP is implemented in Infer.NET in the form of expectation propagation~\cite{Minka:2001:EPA:647235.720257} and is used for the practical part of this report.

\subsection{Approximate inference by sampling}\label{sec:approx_inference}

Another principle of approximative inference is the generation of random samples based on the local probability distributions modelled in the network. The relative frequency of the generated values gives an approximation of the probability. The larger the sample, the more accurate the calculated value.

\begin{itemize}
	\item \textbf{Direct Sampling} generates the samples based on the probability distributions in topological order, i.\,e.\ from parent to child nodes. Let $N$ be the number of all generated samples and $N(\prb{x})$ the number of samples which are consistent with the specific event $\prb{x}$, then the approximated probability of this event amounts to:
	\[\hat{P}(\prb{x}) = \frac{N(\prb{x})}{N} \approx P(\prb{x}).\]
	This naive approach cannot handle evidence.
	\item \textbf{Rejection sampling} can provide approximations to the distributions in the form $\hat{\prb{P}}(x|\prb{e})$ by using only the samples that are consistent with the observed evidence \textbf{e}. As many samples are rejected, this method is quite inefficient. The probability distribution of a variable is approximated as follows:
	\[\hat{\prb{P}}(X|\prb{e}) = \frac{\prb{N}(X,\prb{e})}{N(e)} \approx \prb{P}(X|\prb{e}).\] 
	\item \textbf{Probability weighting}  is more efficient than rejection sampling because it does not generate unused samples and only samples variables that are not evidence variables. Each sample is weighted according to the probability with which it corresponds to the evidence. Samples with a higher weighting are included in the calculation to a greater extent than samples with a lower weighting. Though as the number of evidence variables increases, the performance gets worse.
	\item \textbf{MCMC algorithms}\footnote{MCMC stands for \textit{Markov chain Monte Carlo}. Such a procedure is also called \textbf{Markov chain simulation}. The term \glqq{}Monte Carlo\grqq{} goes back to John von Neumann, who suggested this name in reference to the casino in Monaco, which is located in the district of the same name~\cite{Monte_Carlo}.} do not create each individual sample from scratch, but by applying random changes to the previous sample. The \textit{current state} specifies a value for each variable, and the \textit{next state} is generated by the MCMC algorithm by randomly changing the current state. \textbf{Gibbs sampling} is a widely used MCMC algorithm, which is particularly suitable for Bayesian networks. Thereby the non-evidence variables are set to a randomly determined initial state before they are sampled. The sampling of a variable is based on the known current values of its Markov blanket variables and changing a variable corresponds to a step in the state space and thus to a new sample. Gibbs sampling is implemented in Infer.NET and can be used as an inference algorithm. However, one drawback is that it is very computationally intensive if sufficient accuracy is to be achieved. Moreover, it is difficult to determine when the algorithm converges \textendash\ any additional iteration can potentially worsen the result again.
\end{itemize}

\subsection{Approximate inference with variational message passing}\label{chapter:theory:sec:inference:subsec:VMP}

In models with many continuous variables or discrete variables with many states, the conditional probability distributions between the individual variables become very complex. The high-dimensional sums and integrals that arise when the probabilities are calculated cannot be determined analytically and a numerical solution or estimation by MCMC sampling requires very intensive calculations. For such problems the method of variation inference was developed. It iteratively optimizes both an approximation of the model evidence $P(\prb V)$ and the posteriori distribution $P(\prb H|\prb V)$, where $\prb V$ are the observed (\textit{visible}) variables of the model and $\prb H$ the \textit{hidden} variables.  

The core concept is a decomposition of the logarithmic model evidence:
\[\text{ln }P(\prb V)=\mathcal L(Q) + \text{KL}(Q(\prb H)\parallel P(\prb H|\prb V)).\]
KL$(Q\parallel P)$ is the Kullback-Leibler divergence, a measure of the difference between the true distribution $P$ and the approximated distribution $Q$. $\mathcal L(Q)$ is a lower bound to ln $P(\prb V)$, also called \textit{free energy}, and is defined by:
\[\mathcal L(Q) = \left<\text{ln }P(\prb V)\right>_Q - \text{KL}(Q(\prb H)\parallel P(\prb H|\prb V)),\] where the estimate $\left<\text{ln }P(\prb V)\right>_Q$ is made using $Q$.
By determining the distribution $Q$ which maximizes the $\mathcal L(Q)$ function, the Kullback-Leibler divergence between $Q(\prb H)$ and the exact posteriori distribution $P(\prb H|\prb V)$ is indirectly minimized. So if $Q(\prb H)= P(\prb H|\prb V)$, then $\mathcal L(Q)=\text{ln }P(\prb V)$ applies. $\mathcal L(Q)$ is now optimized by the implication of simplifying restrictions on the functional form of $Q$. First, $Q$ should be a factorizable probability distribution with
\[Q(\prb H) = \prod_i Q_i(H_i)\],
where the individual $Q_i$s approximate a posterior distribution of independent groups of variables $\prb H_i$, while $\mathcal L(Q)$ approximates the log evidence ln $P(\prb V)$. Furthermore, the individual $Q_i$s must be selected from a certain family of distributions, a so-called exponential family. Members of an exponential family can be formed using a natural parameter vector $\phi$, a vector $\prb u$ from a sufficient statistic\footnote{A sufficient statistic is a sample from which the underlying distribution can be derived completely.} and a normalization factor $g$:
\[P(\prb X|\prb Y)=\text{exp}\left[\phi(\prb Y)^\text{T}\prb u(\prb X) + f(\prb X) + g(\prb Y)\right].\]
Within one exponential family, the parameter and statistic vectors each have the same form. The individual factors are efficiently optimized by exchanging the estimates of these parameter and statistics vectors between the nodes in the form of a message passing schema. 

The advantage of VMP is that even very complex models from discrete and continuous distributions can be efficiently approximated. The disadvantage is that the approximation cannot be arbitrarily accurate. For observed discrete variables with more than two states, the approximation does not correspond to the exact distribution due to the functional limitation of the $Q_i$s. In small and purely discrete models, this error may propagate to such an extent that the algorithm produces worse results than other inference algorithms which are computationally still easy to handle and can deliver more accurate approximations with small model sizes and mainly binary variables. How this works in practice is described in section~\ref{chapter:implementation:sec:results:subsec:vmp}.

\section{Prototypical Software Architecture}\label{chapter:Prototyp}

In this chapter, the approaches for the development of a practical agent investigated so far will be presented and evaluated. First of all, the Bayesian network for the diagnosis of headaches developed for this purpose will be outlined. Then the implemented agent and the procedure for translating the external format into the Infer.NET form and the inference with Infer.NET itself are explained.

\subsection{Bayesian Network for the diagnosis of headaches}

\begin{figure}[H]
 \centering
 \makebox[\textwidth][c]{\includegraphics[width=1.2\textwidth]{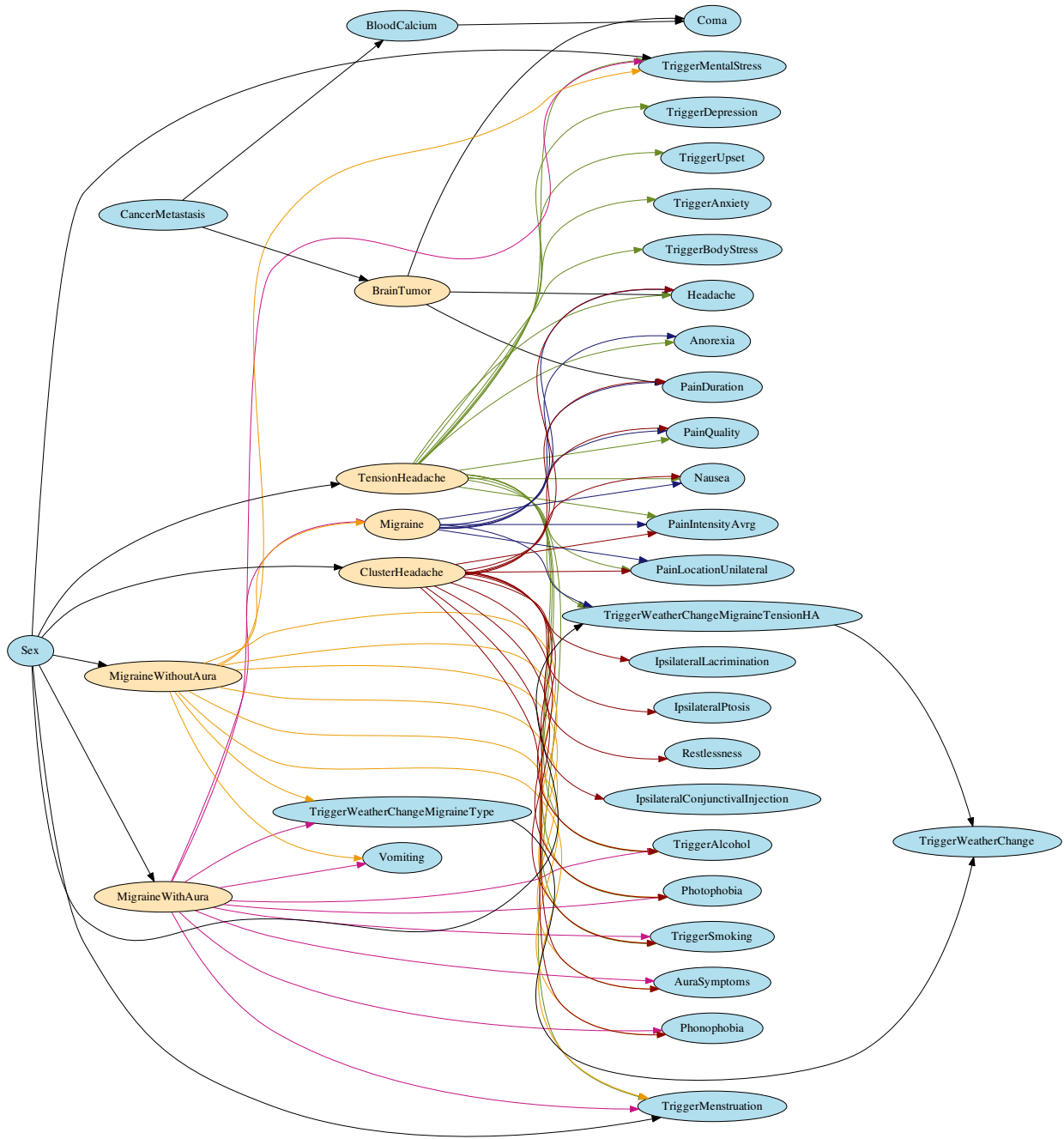}}
 \caption{Bayesian network for headache diagnosis.}\label{fig:full_network}
\end{figure}

In order to test the inference methods, at first a Bayesian network for pain diagnosis was modelled. The model was based on the diagnoses and their hierarchy was mapped in the network structure in such a way that more specific diagnoses were modelled as parent nodes of more general ones. The prototype is limited to an example network for the diagnosis of a brain tumor and three different forms of headache (migraine, tension headache and cluster headache) with the corresponding symptoms (figure~\ref{fig:full_network}). The migraine was preceded by two more specific diagnoses: migraine with aura and migraine without aura.

\subsection{Representing the knowledge base}\label{chapter:implementation:sec:network_representation}

In order to have queries sent to a network via an inference package, they have to be submitted by means of an appropriate representation. Considering the above-mentioned components any discrete Bayesian network can be constructed. Representing a Bayesian network as source code however is rather unsuitable for the construction of a knowledge base by a domain expert. It is not portable, confusing and difficult to process with external tools. Instead, a format which is clearly structured, easy to parse and also easy to edit by domain experts in an editor is desired. Additionally, it should provide a compact representation form for more complex node functions such as the canonical probabilistic models presented in chapter~\ref{sec:ICI models}. Since there already are a number of text formats for Bayesian networks available, the development of a novel format is not necessary here. PropmodelXML was developed e.\,g.\ for the representation of various probabilistic graphical models and is suitable for use in the discourse area. A detailed description of the format can be found in~\cite{Diez00canonicalprobabilistic}. The most important criterion is the full support for ICI models, which is not provided by any other format. On top of that, it is an XML format that is flexible and extensible.

\subsection{Implementing a prototype using Infer.NET}\label{chapter:implementation:sec:infer.net_agent}

In order to test the presented methods for their practicability, a prototypical agent was implemented in C\#. The program is able to parse a network in the PGMX format and can convert it into an Infer.NET model definition (section~\ref{chapter:implementation:sec:infer.net_agent:model_transformation}). The execution of queries on the imported network using Infer.NET is shown in section~\ref{chapter:implementation:sec:infer.net_agent:subsec:inference}. Moreover, by generating sample data sets for and processing them with a specific network, the network can be analysed in more detail. 

\subsubsection{Transferring the external knowledge representation}\label{chapter:implementation:sec:infer.net_agent:model_transformation}

The transformation of the external network representation into a model on which the Infer.NET package can perform inferences is done step by step\footnote{Infer.NET, developed by Microsoft Research, is a library for inference on probabilistic graphical models such as Bayesian networks or Markov random fields. Expectation propagation, variational message passing and Gibbs sampling are inference algorithms that are implented in this library.}: \\

\textbf{1. Parsing} \\
First, the elements of a Bayesian network are parsed from the XML file, converted into equivalent data structures in C\# and combined to form a network structure in which each child node knows its parent nodes. The most important building block here is \texttt{BayesianNode}, which represents a network node. \\

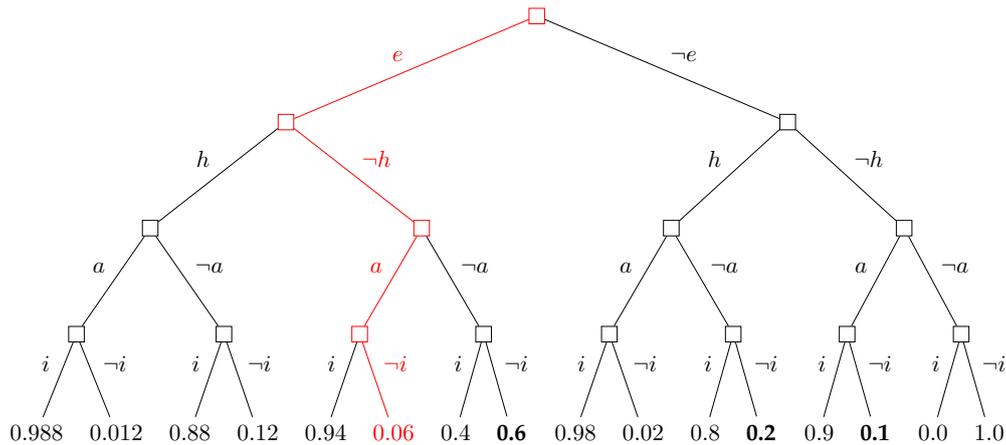
\begin{figure}[t]
\centering
\begin{tikzpicture}[
  scale=0.8,
  every level 0 node/.style={draw},
  every level 1 node/.style={draw},
  every level 2 node/.style={draw},
  every level 3 node/.style={draw},
  every leaf node/.style={draw=none,red},
  sibling distance=5pt,
  level distance=50pt
]
\tikzset{edge from parent/.style={draw, edge from parent path=
    {(\tikzparentnode) -- (\tikzchildnode)}}}
\Tree 
[.\node[red]{};
  \edge[red] node[auto=right] {$e$};
  [.\node[red]{}; 
    \edge node[auto=right] {$h$};
    [.\node{}; 
      \edge node[auto=right] {$a$};
      [.\node{};
        \edge node[auto=right] {$i$};
        [.\node{$0.988$};] 
        \edge node[auto=left] {$\neg i$};
        [.\node{$0.012$};]] 
      \edge node[auto=left] {$\neg a$};
      [.\node{};
        \edge node[auto=right] {$i$};
        [.\node{$0.88$};] 
        \edge node[auto=left] {$\neg i$};
        [.\node{$0.12$};]]]
    \edge[red] node[auto=left] {$\neg h$};
    [.\node[red]{}; 
      \edge[red] node[auto=right] {$a$};
      [.\node[red]{};
        \edge node[auto=right] {$i$};
        [.\node{$0.94$};] 
        \edge[red] node[auto=left] {$\neg i$};
        [.\node[red]{$0.06$};]] 
      \edge node[auto=left] {$\neg a$};
      [.\node{};
        \edge node[auto=right] {$i$};
        [.\node{$0.4$};] 
        \edge node[auto=left] {$\neg i$};
        [.\node{$\textbf{0.6}$};]]]] 
  \edge node[auto=left] {$\neg e$};
  [.\node{}; 
    \edge node[auto=right] {$h$};
    [.\node{}; 
      \edge node[auto=right] {$a$};
      [.\node{};
        \edge node[auto=right] {$i$};
        [.\node{$0.98$};] 
        \edge node[auto=left] {$\neg i$};
        [.\node{$0.02$};]] 
      \edge node[auto=left] {$\neg a$};
      [.\node{};
        \edge node[auto=right] {$i$};
        [.\node{$0.8$};] 
        \edge node[auto=left] {$\neg i$};
        [.\node{$\textbf{0.2}$};]]]
    \edge node[auto=left] {$\neg h$};
    [.\node{}; 
      \edge node[auto=right] {$a$};
      [.\node{};
        \edge node[auto=right] {$i$};
        [.\node{$0.9$};] 
        \edge node[auto=left] {$\neg i$};
        [.\node{$\textbf{0.1}$};]] 
      \edge node[auto=left] {$\neg a$};
      [.\node{};
        \edge node[auto=right] {$i$};
        [.\node{$0.0$};] 
        \edge node[auto=left] {$\neg i$};
        [.\node{$1.0$};]]]]]
\end{tikzpicture}
\caption{Conditional probability table as a tree-like structure for the noisy OR example \glqq{} Heart attack ($i$) with possible causes endocarditis ($e$), hypertension ($h$) or arteriosclerosis ($a$)\grqq. The path through the tree which is marked in red gives the probability $P(\neg i|e \wedge \neg h \wedge a) = 0.06$. The bold-typed probabilities are the inhibitor parameters of the noisy OR.}\label{CPT-Tree Herzinfarkt}
\end{figure}

\textbf{2. Construction of conditional probability trees} \\
If Infer.NET-compatible C\# code is to be generated from the probabilities of a node (represented by so-called \textit{potentials} in the PGMX format) using this structure, the nested case distinctions of a conditional probability distribution can best be generated recursively from a tree. At the same time, a tree is a very efficient form of representing CPTs. Figure~\ref{CPT-Tree Herzinfarkt} depicts the CPT from table~\ref{tab:noisy-or_coronary_example} in a tree structure. Each branching depth below the tree's root represents a variable column of the CPT and thus each node level (except the root) stands for a variable in the conditional probability distribution. The last branch illustrates the conditional variable (the child node). The last branch represents the conditional variable (the child node). The edges of the tree are labelled with the respective value assignments of the underlying variable, so that each path through the tree -- from the root to a leaf -- results in a unique combination of assignments and therefore represents exactly one row in the CPT. The leaves are labelled with the corresponding conditional probabilities and it is very easy to navigate through the resulting data structure.

The probabilities from the external representation are converted into these kinds of conditional probability trees and assigned to their network nodes. With ICI models (noisy OR or similar), most of the conditional probabilities in the leaves are calculated using the corresponding formulas from section~\ref{sec:ICI models}. In order to have each tree node related to the variable to whose column it belongs, it gets a reference to the corresponding node of the network when the tree is created.\\

\textbf{3. Generating the Infer.NET model} \\
Finally, corresponding Infer.NET variables are created for all network nodes, with each node receiving a reference to its associated Infer.NET variable. Now the Infer.NET variables can be linked using the conditional probability trees. This is done by recursively descending into the CPT tree and generating a case distinction within a \texttt{using} statement for the Infer.NET variable associated with the current tree level. Once the leaf level is reached, the probability distribution defined in the leaves is assigned to the associated Infer.NET variable.

\subsubsection{Inference with the agent}
\enlargethispage{-\baselineskip}\label{chapter:implementation:sec:infer.net_agent:subsec:inference}
The following paragraph explains how the implemented prototype uses Infer.NET to perform queries on the constructed model definition. It therefore describes how the Infer.NET inference engine works.  

\begin{figure}[H]
\centering
\includegraphics[width=1\textwidth]
{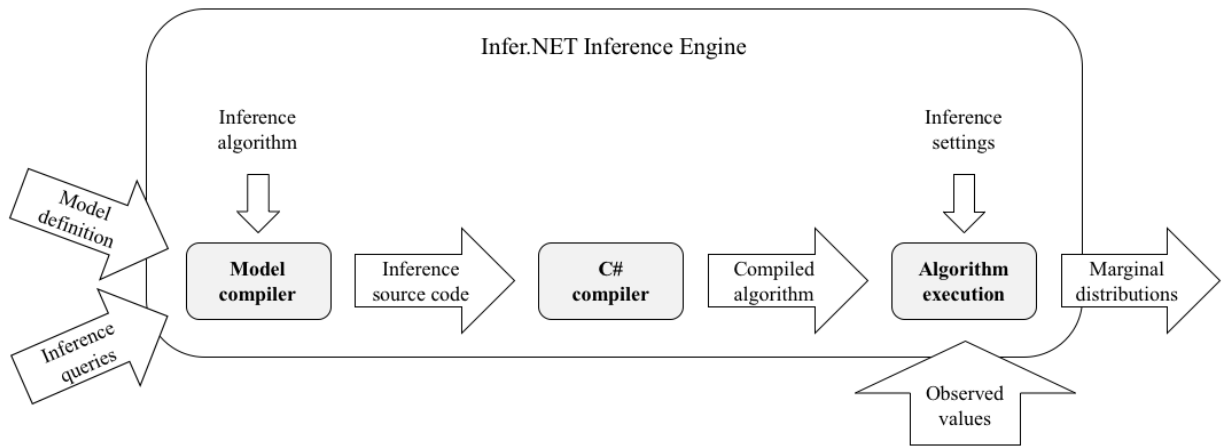}
\caption{Working scheme of Infer.NET~\cite{InferNET18}\label{fig:how_infer.net_works}.}
\end{figure}

\begin{enumerate}
\item First, a model definition is created, which is described using the modeling API. This task is handed to the transformation algorithm introduced in section~\ref{chapter:implementation:sec:infer.net_agent:model_transformation}. Moreover, the inference algorithm is defined and the inference queries to the model are declared by specifying the observed variables and the query variables.
\item The model is passed to the model compiler, which generates optimized C\# source code that makes it possible to query the model based on the specified inference algorithm. The generated source code is written to a file and can later be used directly.
\item The source code is compiled using the C\# compiler. It is possible to execute this step manually in order to control how the inference is performed. The easiest way is to run it automatically with the help of the \texttt{Infer} method.
\item With a set of concrete values for the observed variables, the inference engine executes the compiled algorithm. This can be repeated for different values of the observed variables without recompiling the algorithm. Which variables are observed, however, is given by a fixed definition for a certain compilation.
\end{enumerate}
Steps 2 to 4 are initiated by calling the \texttt{Infer} method. Infer.NET decides if the algorithm has to be recompiled when the \texttt{Infer} method is called again.

\section{Evaluation}\label{chapter:implementation:sec:test}

\subsection{Evaluation of the network quality using real data}\label{chapter:implementation:sec:test:subsec:ebaybbn}

In order to assess how well the network classifies the modelled diagnoses, some tests based on real anamnesis data were carried out in the previous work~\cite{beleg}. The procedure is presented here again to compare it with the improved method in section~\ref{chapter:implementation:sec:test:subsec:infer.net:subsubsec:sample_creation}. Data sets consisting of diagnoses and anamneses were selected and used to manually generate queries to the model. With the implemented network, probabilities for the different diagnoses were calculated from these queries and compared with the diagnoses that were obtained by physicians and are included in the data sets. High probabilities were expected for the diagnoses made and significantly lower probabilities for the diagnoses not made. 

A query consists of a set of evidence variables and the associated observed values. The anamnesis data of each data set were evaluated manually. If a feature modelled in the network was sufficiently described in the anamnesis, an evidence variable was added to the query and assigned a value that matched the anamnesis description (table~\ref{Testabfragen}).

\begin{table}[t]
	\centering
	\captionsetup{width=1.0\textwidth}
	\scriptsize
	\renewcommand{\arraystretch}{1.3} 
	\begin{tabular}{lllllllll}
		\toprule
		\textbf{Characteristic} & \textbf{P1} & \textbf{P2} & \textbf{P3} & \textbf{P4} & \textbf{P5} & \textbf{P6} & \textbf{P7} & \textbf{P8} \\ \toprule 
		Anorexia     & True & False & & False & & & & boolean   \\  \midrule
		AuraSymptoms & False & & False & True & True & & & False \\  \midrule
		Headache & True & True & True & True & True & True & True & True \\  \midrule
		IpsilateralConjunctivalInjection & False & False & False & False & & & True & \\  \midrule
		IpsilateralLacrimination & True & & False & False & & & True & \\  \midrule
		IpsilateralPtosis & False & False & & False & & & & \\  \midrule
		Nausea & True & False & True & False & & & & True \\  \midrule
		PainDuration & & & Days & Days & Years & Years & Hours & Hours \\  \midrule
		PainIntensityAvrg & 3 & 2 & 8 & 3 & 7 & 7 & 7 & 5 \\  \midrule
		PainLocationUnilateral & False & & True & False & & & True & \\  \midrule
		PainQuality & Pulsating & Stabbing & Stabbing & & Other & Stabbing & Other & Stabbing \\  \midrule
		Phonophobia & True & True & True & True & & & & \\  \midrule
		Photophobia & True & True & & False & & & & \\  \midrule
		Restlessness & False & False & True & & & & & \\  \midrule 
		Sex & Female & Male & Female & Female & Male & Female & Male & Male \\  \midrule
		TriggerAlcohol & True & & & True & True & & & False \\  \midrule
		TriggerAnxietyDepression & & & True & & False & & & False \\  \midrule
		TriggerBodyStress & True & True & False & & True & True & & False \\  \midrule
		TriggerMenstruation & & & True & True & & & & \\  \midrule
		TriggerMentalStress & True & & True & True & True & True & & False \\  \midrule
		TriggerSmoking & & & & & & & & False \\  \midrule
		TriggerUpset & True & & & True & & True & & False \\  \midrule
		TriggerWeather & True & True & & True & & True & & False \\  \midrule
		Vomiting & False & False & True & False & & & & \\  \bottomrule 
	\end{tabular}
	\caption{Test queries for eight patients.}\label{Testabfragen}
\end{table}

For each query, the network was used to calculate the conditional probability distributions of all nodes. Most importantly, the probabilities of the diagnoses are listed in table~\ref{Testergebnisse}.

\begin{table}[H]
	\small
	\centering
	\captionsetup{width=1.0\textwidth}
	\begin{tabular}{lllllll}
		\toprule
		\textbf{Patients} & \textbf{Diagnoses} & \textbf{CHA} & \textbf{Migraine} & \textbf{MA} & \textbf{MO} & \textbf{TTHA} \\ \toprule 
		Patient 1 & MO 		&   0,003 & 0,9997 & 0,0172 & 0,9847 & 0,9907 \\  
		Patient 2 & MO 		&   0,0086 & 0,9914 & 0,5904 & 0,4073 & 0,3892 \\  
		Patient 3 & MO		&   0,0011 & 1 & 0,018 & 0,992 & 0,0342 \\  
		Patient 4 & MA		&   0,0003 & 0,9997 & 0,9934 & 0,02 & 0,9289 \\  
		Patient 5 & TTHA	&   0,0354 & 0,9768 & 0,9679  & 0,0817 & 0,5799 \\  
		Patient 6 & TTHA	&   0,0006 & 0,1953 & 0,068 & 0,1376 & 0,9948 \\  
		Patient 7 & CHA		&   0,9979 & 0,0399 & 0,0242 & 0,0161 & 0,008 \\  
		Patient 8 & CHA		&   0,1022 & 0,8951 & 0,0936 & 0,8022 & 0,0048 \\ \bottomrule
	\end{tabular}
	\caption{Test results\@: Diagnoses and expected diagnoses; CHA\@: cluster headache; MA\@: migraine with aura; MO\@: migraine without aura; TTHA\@: tension-type headache.}\label{Testergebnisse}
\end{table}

The results of the tests show that in six out of eight cases the diagnosis made by the physician is predicted by the network with a high probability, although only a small part of the variables relevant for the domain were modelled in the network. Also, the diagnostic probability for the other diagnoses is usually very low. The \glqq{}false\grqq{} classifications can be explained, too:

\begin{itemize}
	\item Patient 1 and 5: The queries contain many symptoms that indicate both diagnoses. Various characteristics in the anamnesis data which make the tension headache distinguishable from the migraine, such as pain localisation and pain perception, but also the temporal course, are not yet modelled in sufficient detail in the network. In addition, symptoms of migraine and tension headache often occur in parallel. This is described as a combination headache. Patient 5 also reported symptoms suggesting an aura, which is why migraine is given more weight than tension headache. He is also listed as a migraine sufferer in the database.
	\item Patient 2: When evaluating the anamnesis, it was difficult to decide whether aura symptoms were present or not based on the data at hand. The network therefore correctly classifies a migraine, but cannot decide exactly which specific type of migraine it is.
	\item Patient 8: Even for an expert it is not easy to correctly distinguish cluster headache from migraine. This is why in the literature usually an explicit reference to a number of criteria for differential diagnosis can be found. Here, none of the ipsilateral symptoms typical of cluster headache, such as conjunctivitis, lacrimation and ptosis were recorded in the anamnesis. Nevertheless, the network classifies cluster headache with a probability almost three times higher than in the other cases where cluster headache was not present.
\end{itemize}

When evaluating the anamnesis data, the general phenomenon could be observed that many patients come to the physician with a fixed notion of a diagnosis. Thus patient 1 mentioned his migraine several times in the anamnesis, so that a certain bias of the physician with respect to the diagnosis cannot be excluded and perhaps a possibly existing tension headache has therfefore not been diagnosed.

Though the tests provide interesting results, the sample size for the evaluation of the network with eight patients is very small. What is more, the quality of the available patient data is insufficient. On the one hand, the anamnesis characteristics are strongly over-specified, while on the other hand they are only insufficiently filled in. Also, for certain diagnoses there were very few or no example patients at all. 

\subsection{Evaluation of the network quality using generated example data}\label{chapter:implementation:sec:test:subsec:infer.net:subsubsec:sample_creation}

In order to test the classification quality of the prototype (section~\ref{chapter:implementation:sec:infer.net_agent}) and to compare it with the prototype of the previous work~\cite{beleg}, a new approach was taken. With the help of the network itself, a priori distributions were determined for the characteristics observed in an anamnesis and example data sets were generated on the basis of these distributions. This method of determining the quality of a network has already been used successfully in the fault diagnosis of motor vehicles~\cite{CarDiagnosisEvaluation}. 

First, suitable examples are generated for each diagnosis, as described below.
\begin{enumerate}
 \item A query with the following properties is created:
 \begin{itemize}
  \item The desired diagnosis is an observed variable with the value \textit{true}.
  \item All other diagnoses are observed variables with the value \textit{false}.
  \item All characteristics observed during diagnostics (e.\,g.\ symptoms) are query variables.
 \end{itemize}
 \item For all characteristics to be observed, the conditional probabilities under the given diagnosis are determined by inference on the query that has just been generated.
 \item An example is generated by randomly choosing the value given the just determined probability distribution for every characteristic to be observed. This step is repeated until the desired number of examples has been generated.
 \item Queries are constructed with the example values as observed variables and all diagnoses as query variables, and the diagnostic probabilities are determined by inference.
 \item The expected diagnosis used to generate the example is compared with the diagnosis for which the highest probability was determined in the previous step.
\end{enumerate}

The sampling was implemented for both prototypes, since the old network partly provides different distributions of the characteristics than the new revised one. Thus, it was also possible to test how robustly the new prototype and its Bayesian network, which is more specific in differentiating its characteristics, deal with examples generated by the old prototype under a less specific characteristic differentiation.

With every algorithm (junction tree belief propagation (JTBP) on the basis of the old network, expectation propagation (EP) and variational message passing (VMP) on the basis of the new network) 200 examples per diagnosis were generated and intermediately stored in a database. With each of these algorithms the diagnostic probabilities for all examples were then calculated and written to the database for evaluation.

To assess the quality of the inference, the proportion of correctly classified examples was determined for each diagnosis and each algorithm (table~\ref{tab:results:correct_classifications}).

\begin{table}[H]
\centering
\small
{\renewcommand{\arraystretch}{1.0}
\begin{tabular}{l l r r r}
	\toprule
&Sampling- & \multicolumn{3}{c}{Inference algorithms}\\
Diagnosis (200 examples each)& algorithm & JTBP  & EP     & VMP    \\ \toprule
BrainTumor & JTBP & 100 & 99 & 99 \\
ClusterHeadache & JTBP & 99 & 98 & 99 \\
MigraineWithAura & JTBP & 97 & 97 & 33 \\
MigraineWithoutAura & JTBP & 100 & 100 & 96 \\
TensionHeadache & JTBP & 97 & 98 & 96 \\ \midrule
BrainTumor & EP & 100 & 100 & 100 \\
ClusterHeadache & EP & 99 & 100 & 100 \\
MigraineWithAura & EP & 93 & 93 & 38 \\
MigraineWithoutAura & EP & 99 & 99 & 94 \\
TensionHeadache & EP & 98 & 99 & 99 \\ \midrule
BrainTumor & VMP & 100 & 100 & 100 \\
ClusterHeadache & VMP & 98 & 98 & 100 \\
MigraineWithAura & VMP & 97 & 97 & 39 \\
MigraineWithoutAura & VMP & 100 & 100 & 91 \\
TensionHeadache & VMP & 95 & 96 & 95 \\ \bottomrule
\end{tabular}}
\caption{Percentages of correctly classified examples.}\label{tab:results:correct_classifications}
\end{table}

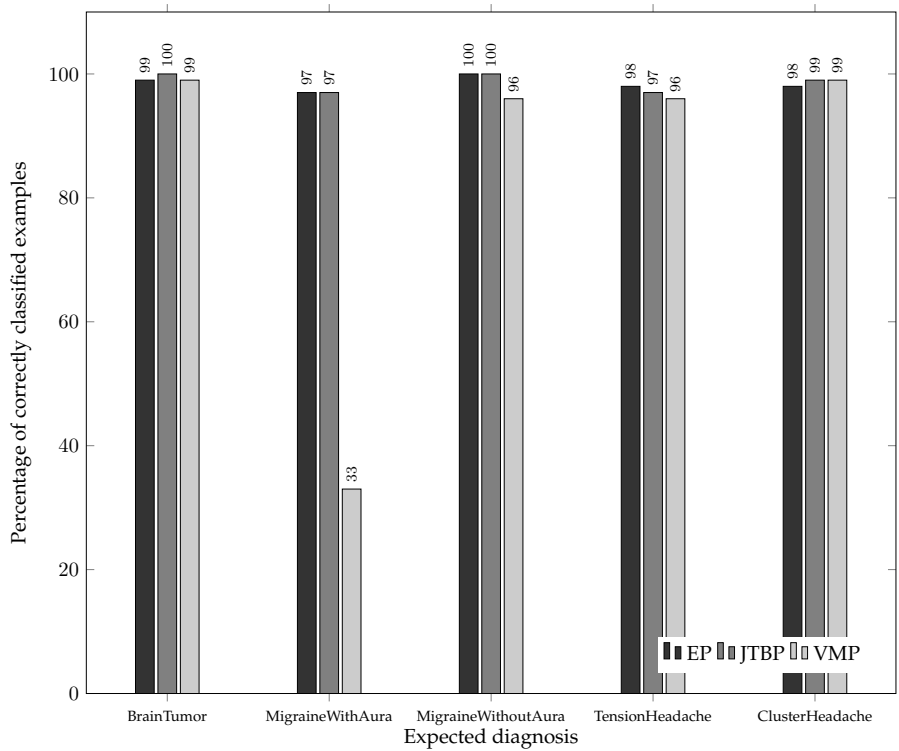
\begin{figure}[H]
	\pgfplotstableread{
		0 0.99 1 0.99
		1 0.97 0.97 0.33
		2 1 1 0.96
		3 0.98 0.97 0.96
		4 0.98 0.99 0.99
	} \dataset{}
	\centering
	\begin{tikzpicture}[scale=0.7]
	\begin{axis}[
	ybar,
	xlabel=Expected diagnosis,
	ylabel=Percentage of correctly classified examples,
	xmin=-0.5,
	xmax=4.5,
	ymin=0,
	ymax=110,
	width=1.05\linewidth,
	xtick={0,1,2,3,4},
	ytick={0,20,40,60,80,100},
	xticklabels = {
		\strut BrainTumor,
		\strut MigraineWithAura,
		\strut MigraineWithoutAura,
		\strut TensionHeadache,
		\strut ClusterHeadache},
	x tick label style={font=\scriptsize},
	every node near coord/.append style={
		anchor=west,
		rotate=90,
		font=\scriptsize},
	legend entries={EP, JTBP, VMP},
	legend columns=3,
	legend style={draw=none,nodes={inner sep=3pt}},
	legend pos=south east,
	]
	\addplot[draw=black, fill=black!80, nodes near coords] table[x index=0, y expr=\thisrowno{1}*100] \dataset;
	\addplot[draw=black, fill=black!50, nodes near coords] table[x index=0, y expr=\thisrowno{2}*100] \dataset;
	\addplot[draw=black, fill=black!20, nodes near coords] table[x index=0, y expr=\thisrowno{3}*100] \dataset;
	\end{axis}
	\end{tikzpicture}
	\caption{Percentages of correctly classified examples by inference with the three algorithms for the examples generated with JTBP (table~\ref{tab:results:correct_classifications}).}
	\label{fig:results:correct_classifications1}
\end{figure}

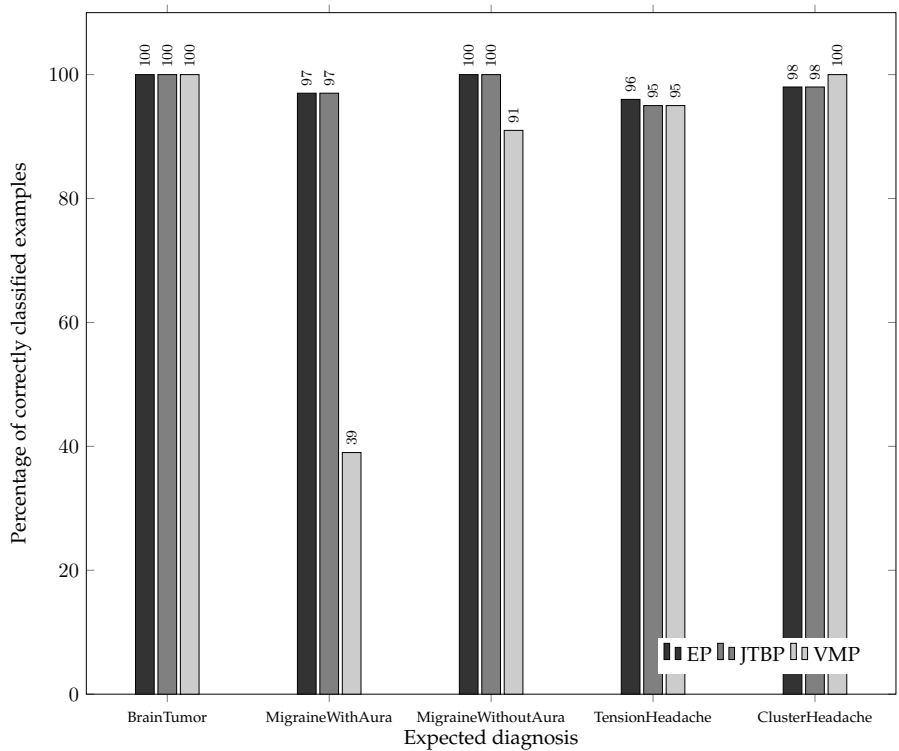
\begin{figure}[H]
	\pgfplotstableread{
		0 1 1 1
		1 0.97 0.97 0.39
		2 1 1 0.91
		3 0.96 0.95 0.95
		4 0.98 0.98 1
	} \dataset{}
	\centering
	\begin{tikzpicture}[scale=0.7]
	\begin{axis}[
	ybar,
	xlabel=Expected diagnosis,
	ylabel=Percentage of correctly classified examples,
	xmin=-0.5,
	xmax=4.5,
	ymin=0,
	ymax=110,
	width=1.05\linewidth,
	xtick={0,1,2,3,4},
	ytick={0,20,40,60,80,100},
	xticklabels = {
		\strut BrainTumor,
		\strut MigraineWithAura,
		\strut MigraineWithoutAura,
		\strut TensionHeadache,
		\strut ClusterHeadache},
	x tick label style={font=\scriptsize},
	every node near coord/.append style={
		anchor=west,
		rotate=90,
		font=\scriptsize},
	legend entries={EP, JTBP, VMP},
	legend columns=3,
	legend style={draw=none,nodes={inner sep=3pt}},
	legend pos=south east,
	]
	\addplot[draw=black, fill=black!80, nodes near coords] table[x index=0, y expr=\thisrowno{1}*100] \dataset;
	\addplot[draw=black, fill=black!50, nodes near coords] table[x index=0, y expr=\thisrowno{2}*100] \dataset;
	\addplot[draw=black, fill=black!20, nodes near coords] table[x index=0, y expr=\thisrowno{3}*100] \dataset;
	\end{axis}
	\end{tikzpicture}
	\caption{Percentages of correctly classified examples by inference with the three algorithms for the examples generated with VMP (table~\ref{tab:results:correct_classifications}).}
	\label{fig:results:correct_classifications2}
\end{figure}

In three experiments with join tree belief propagation (JTBP), expectation propagation (EP) and variational message passing (VMP) 200 examples per diagnosis were generated from the headache networks (\cite{beleg} and figure~\ref{fig:full_network}) and classified with all three algorithms. Table~\ref{tab:results:correct_classifications} shows the proportion of correctly classified examples, and figures \ref{fig:results:correct_classifications} to \ref{fig:results:correct_classifications2} further illustrate the result.

\begin{figure}[H]
	\pgfplotstableread{
		0 1 1 1
		1 0.93 0.93 0.38
		2 0.99 0.99 0.94
		3 0.99 0.98 0.99
		4 1 0.99 1
	} \dataset{}
	\centering
	\begin{tikzpicture}[scale=0.7]
	\begin{axis}[
	ybar,
	xlabel=Expected diagnosis,
	ylabel=Percentage of correctly classified examples,
	xmin=-0.5,
	xmax=4.5,
	ymin=0,
	ymax=110,
	width=1.05\linewidth,
	xtick={0,1,2,3,4},
	ytick={0,20,40,60,80,100},
	xticklabels = {
		\strut Brain,
		\strut MigraineWithAura,
		\strut MigraineWithoutAura,
		\strut TensionHeadache,
		\strut ClusterHeadache},
	x tick label style={font=\scriptsize},
	every node near coord/.append style={
		anchor=west,
		rotate=90,
		font=\scriptsize},
	legend entries={EP, JTBP, VMP},
	legend columns=3,
	legend style={draw=none,nodes={inner sep=3pt}},
	legend pos=south east,
	]
	\addplot[draw=black, fill=black!80, nodes near coords] table[x index=0, y expr=\thisrowno{1}*100] \dataset;
	\addplot[draw=black, fill=black!50, nodes near coords] table[x index=0, y expr=\thisrowno{2}*100] \dataset;
	\addplot[draw=black, fill=black!20, nodes near coords] table[x index=0, y expr=\thisrowno{3}*100] \dataset;
	\end{axis}
	\end{tikzpicture}
	\caption{Percentages of correctly classified examples by inference with the three algorithms for the examples generated with EP (tabelle~\ref{tab:results:correct_classifications}).}
	\label{fig:results:correct_classifications}
\end{figure}

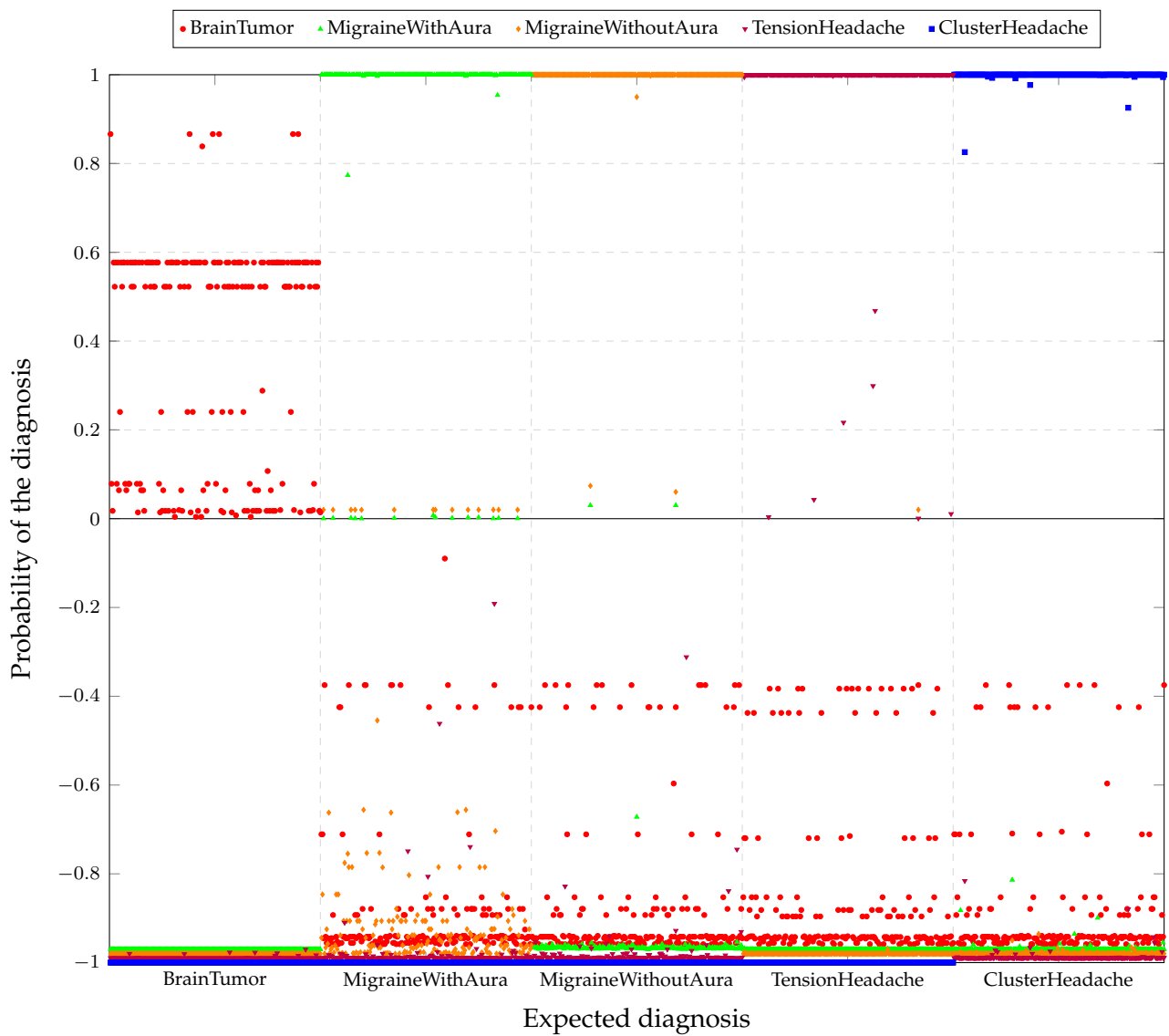
\begin{figure}[H]
	\makebox[\textwidth][c]{
		\centering
		\begin{tikzpicture}[scale=1.0]
		\pgfplotsset{every axis legend/.append style={
				at={(0.5,1.03)},
				anchor=south}}
		\begin{axis}[
		xlabel=Expected diagnosis,
		ylabel=Probability of the diagnosis,
		legend style={font=\scriptsize,/tikz/every even column/.append style={column sep=0.28cm}},
		legend columns=5,
		xtick={100,300,500,700,900},
		xticklabels={BrainTumor,MigraineWithAura,MigraineWithoutAura,TensionHeadache,ClusterHeadache},
		x tick label style={font=\scriptsize},
		y tick label style={font=\scriptsize},
		width=1.05\linewidth, 
		minor xtick={200,400,600,800},
		minor ytick={0.2,0.4,0.6,0.8},
		grid=minor,
		grid style={dashed,gray!30},
		xmin=0, 
		xmax=1000,
		ymin=-1,
		ymax=1,
		]
		\addplot[black, no markers, forget plot] coordinates {(0,0) (1000,0)};
		\addplot[red, only marks, mark=*, mark options={mark size=1}] 
		table[x=x,y expr=\thisrow{BrainTumorF}*-1+0.04,col sep=comma] 
		{chapters/tables/JoinedResultProbsEPSamplesEPInferredWithInf.csv};
		\addplot[green, only marks, mark=triangle*, mark options={mark size=1}] 
		table[x=x,y expr=\thisrow{MigraineWithAuraF}*-1+0.03,col sep=comma] 
		{chapters/tables/JoinedResultProbsEPSamplesEPInferredWithInf.csv};
		\addplot[orange, only marks, mark=diamond*, mark options={mark size=1}] 
		table[x=x,y expr=\thisrow{MigraineWithoutAuraF}*-1+0.02,col sep=comma] 
		{chapters/tables/JoinedResultProbsEPSamplesEPInferredWithInf.csv};
		\addplot[purple, only marks, mark=triangle*, mark options={mark size=1,rotate=180}] 
		table[x=x,y expr=\thisrow{TensionHeadacheF}*-1+0.01,col sep=comma] 
		{chapters/tables/JoinedResultProbsEPSamplesEPInferredWithInf.csv};
		\addplot[blue, only marks, mark=square*, mark options={mark size=1}] 
		table[x=x,y expr=\thisrow{ClusterHeadacheF}*-1,col sep=comma] 
		{chapters/tables/JoinedResultProbsEPSamplesEPInferredWithInf.csv};
		\addplot[blue, only marks, mark=square*, mark options={mark size=1}] 
		table[x=x,y=ClusterHeadacheT,col sep=comma] {chapters/tables/JoinedResultProbsEPSamplesEPInferredWithInf.csv};
		\addplot[purple, only marks, mark=triangle*, mark options={mark size=1,rotate=180}] 
		table[x=x,y=TensionHeadacheT,col sep=comma] 
		{chapters/tables/JoinedResultProbsEPSamplesEPInferredWithInf.csv};
		\addplot[orange, only marks, mark=diamond*, mark options={mark size=1}] 
		table[x=x,y=MigraineWithoutAuraT,col sep=comma] 
		{chapters/tables/JoinedResultProbsEPSamplesEPInferredWithInf.csv};
		\addplot[green, only marks, mark=triangle*, mark options={mark size=1}] 
		table[x=x,y=MigraineWithAuraT,col sep=comma] 
		{chapters/tables/JoinedResultProbsEPSamplesEPInferredWithInf.csv};
		\addplot[red, only marks, mark=*, mark options={mark size=1}] 
		table[x=x,y=BrainTumorT,col sep=comma] 
		{chapters/tables/JoinedResultProbsEPSamplesEPInferredWithInf.csv};
		\legend{BrainTumor,MigraineWithAura,MigraineWithoutAura,TensionHeadache,ClusterHeadache}
		\end{axis}
		\end{tikzpicture}}
	\caption{Results of the inference with expectation propagation for the examples generated with EP.}\label{fig:results:full_plot:ep_samples:ep_inference}
\end{figure}

\begin{figure}[H]
	\makebox[\textwidth][c]{
		\centering
		\begin{tikzpicture}[scale=1.0]
		\pgfplotsset{every axis legend/.append style={
				at={(0.5,1.03)},
				anchor=south}}
		\begin{axis}[
		xlabel=Expected diagnosis,
		ylabel=Probability of the diagnosis,
		legend style={font=\scriptsize,/tikz/every even column/.append style={column sep=0.28cm}},
		legend columns=5,
		xtick={100,300,500,700,900},
		xticklabels={BrainTumor,MigraineWithAura,MigraineWithoutAura,TensionHeadache,ClusterHeadache},
		x tick label style={font=\scriptsize},
		y tick label style={font=\scriptsize},
		width=1.05\linewidth, 
		minor xtick={200,400,600,800},
		minor ytick={0.2,0.4,0.6,0.8},
		grid=minor,
		grid style={dashed,gray!30},
		xmin=0, 
		xmax=1000,
		ymin=-1,
		ymax=1,
		]
		\addplot[black, no markers, forget plot] coordinates {(0,0) (1000,0)};
		\addplot[red, only marks, mark=*, mark options={mark size=1}] 
		table[x expr=\thisrow{x}-1000,y expr=\thisrow{BrainTumorF}*-1+0.04,col sep=comma] 
		{chapters/tables/JoinedResultProbsEPSamplesVMPInferredWithInf.csv};
		\addplot[green, only marks, mark=triangle*, mark options={mark size=1}] 
		table[x expr=\thisrow{x}-1000,y expr=\thisrow{MigraineWithAuraF}*-1+0.03,col sep=comma] 
		{chapters/tables/JoinedResultProbsEPSamplesVMPInferredWithInf.csv};
		\addplot[orange, only marks, mark=diamond*, mark options={mark size=1}] 
		table[x expr=\thisrow{x}-1000,y expr=\thisrow{MigraineWithoutAuraF}*-1+0.02,col sep=comma] 
		{chapters/tables/JoinedResultProbsEPSamplesVMPInferredWithInf.csv};
		\addplot[purple, only marks, mark=triangle*, mark options={mark size=1,rotate=180}] 
		table[x expr=\thisrow{x}-1000,y expr=\thisrow{TensionHeadacheF}*-1+0.01,col sep=comma] 
		{chapters/tables/JoinedResultProbsEPSamplesVMPInferredWithInf.csv};
		\addplot[blue, only marks, mark=square*, mark options={mark size=1}] 
		table[x expr=\thisrow{x}-1000,y expr=\thisrow{ClusterHeadacheF}*-1,col sep=comma] 
		{chapters/tables/JoinedResultProbsEPSamplesVMPInferredWithInf.csv};
		\addplot[blue, only marks, mark=square*, mark options={mark size=1}] 
		table[x expr=\thisrow{x}-1000,y=ClusterHeadacheT,col sep=comma]
		{chapters/tables/JoinedResultProbsEPSamplesVMPInferredWithInf.csv};
		\addplot[purple, only marks, mark=triangle*, mark options={mark size=1,rotate=180}] 
		table[x expr=\thisrow{x}-1000,y=TensionHeadacheT,col sep=comma] 
		{chapters/tables/JoinedResultProbsEPSamplesVMPInferredWithInf.csv};
		\addplot[orange, only marks, mark=diamond*, mark options={mark size=1}] 
		table[x expr=\thisrow{x}-1000,y=MigraineWithoutAuraT,col sep=comma] 
		{chapters/tables/JoinedResultProbsEPSamplesVMPInferredWithInf.csv};
		\addplot[green, only marks, mark=triangle*, mark options={mark size=1}] 
		table[x expr=\thisrow{x}-1000,y=MigraineWithAuraT,col sep=comma] 
		{chapters/tables/JoinedResultProbsEPSamplesVMPInferredWithInf.csv};
		\addplot[red, only marks, mark=*, mark options={mark size=1}] 
		table[x expr=\thisrow{x}-1000,y=BrainTumorT,col sep=comma] 
		{chapters/tables/JoinedResultProbsEPSamplesVMPInferredWithInf.csv};
		\legend{BrainTumor,MigraineWithAura,MigraineWithoutAura,TensionHeadache,ClusterHeadache}
		\end{axis}
		\end{tikzpicture}}
	\caption{Results of the inference with variational message passing for the examples generated with EP.}\label{fig:results:full_plot:ep_samples:vmp_inference}
\end{figure}
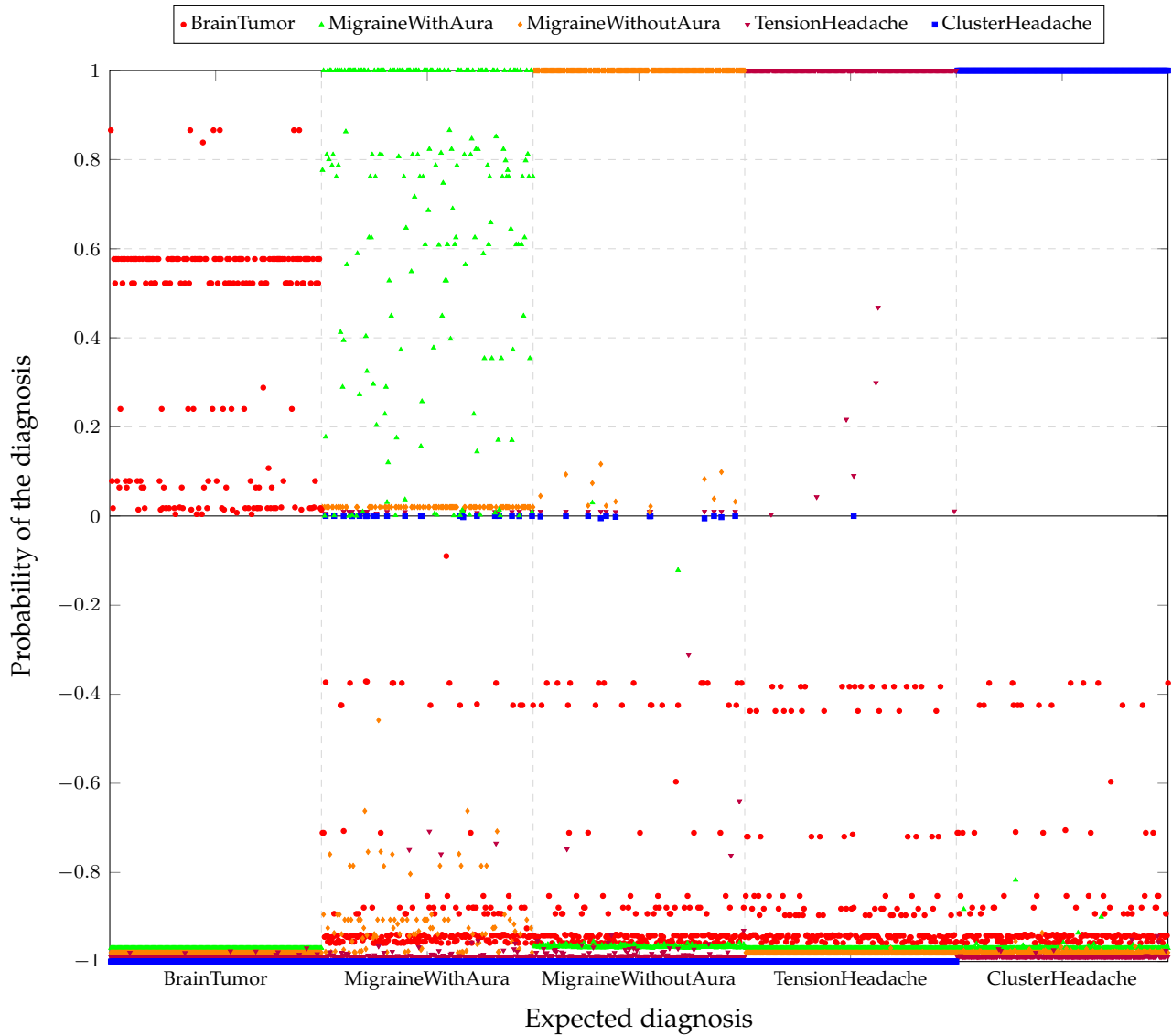

A complete depiction of the diagnostic probabilities determined for the examples is shown in figures~
\ref{fig:results:full_plot:ep_samples:ep_inference} (generated with EP, inference by EP),~
\ref{fig:results:full_plot:ep_samples:jtbp_inference} (generated with EP, inference by JTBP) and~\ref{fig:results:full_plot:ep_samples:vmp_inference} (generated with EP, inference by VMP).

The probability for the occurrence of the expected diagnosis is indicated by a marker in the positive range of the y-axis from 0 to 1. For the unanticipated diagnoses, the probability that they will not occur is given by the markers in the negative range of the y-axis from -1 to 0. The marker on the x-axis shows which diagnosis was expected. The negative values were shifted by up to 0.04 to make the markers visible for every diagnosis. 

What is more, the misclassifications were investigated further. It was determined to what extent the probability of the expected diagnosis differs from the probability of the classified diagnosis. In order to do so, the diagnostic probabilities for misclassified examples generated with expectation propagation were calculated. The probability distributions of the misclassifications for the examples generated with the other two algorithms show no statistically significant differences.

\begin{figure}[H]
	\makebox[\textwidth][c]{
		\centering
		\begin{tikzpicture}[scale=1.0]
		\pgfplotsset{every axis legend/.append style={
				at={(0.5,1.03)},
				anchor=south}}
		\begin{axis}[
		xlabel=Expected diagnosis,
		ylabel=Probability of the diagnosis,
		legend style={font=\scriptsize,/tikz/every even column/.append style={column sep=0.28cm}},
		legend columns=5,
		xtick={100,300,500,700,900},
		xticklabels={BrainTumor,MigraineWithAura,MigraineWithoutAura,TensionHeadache,ClusterHeadache},
		x tick label style={font=\scriptsize},
		y tick label style={font=\scriptsize},
		width=1.05\linewidth, 
		minor xtick={200,400,600,800},
		minor ytick={0.2,0.4,0.6,0.8},
		grid=minor,
		grid style={dashed,gray!30},
		xmin=0, 
		xmax=1000,
		ymin=-1,
		ymax=1,
		]
		\addplot[black, no markers, forget plot] coordinates {(0,0) (1000,0)};
		\addplot[red, only marks, mark=*, mark options={mark size=1}] 
		table[x expr=\thisrow{x}-2000,y expr=\thisrow{BrainTumorF}*-1+0.04,col sep=comma] 
		{chapters/tables/JoinedResultProbsEPSamplesJTBPInferredWithInf.csv};
		\addplot[green, only marks, mark=triangle*, mark options={mark size=1}] 
		table[x expr=\thisrow{x}-2000,y expr=\thisrow{MigraineWithAuraF}*-1+0.03,col sep=comma] 
		{chapters/tables/JoinedResultProbsEPSamplesJTBPInferredWithInf.csv};
		\addplot[orange, only marks, mark=diamond*, mark options={mark size=1}] 
		table[x expr=\thisrow{x}-2000,y expr=\thisrow{MigraineWithoutAuraF}*-1+0.02,col sep=comma] 
		{chapters/tables/JoinedResultProbsEPSamplesJTBPInferredWithInf.csv};
		\addplot[purple, only marks, mark=triangle*, mark options={mark size=1,rotate=180}] 
		table[x expr=\thisrow{x}-2000,y expr=\thisrow{TensionHeadacheF}*-1+0.01,col sep=comma] 
		{chapters/tables/JoinedResultProbsEPSamplesJTBPInferredWithInf.csv};
		\addplot[blue, only marks, mark=square*, mark options={mark size=1}] 
		table[x expr=\thisrow{x}-2000,y expr=\thisrow{ClusterHeadacheF}*-1,col sep=comma] 
		{chapters/tables/JoinedResultProbsEPSamplesJTBPInferredWithInf.csv};
		\addplot[blue, only marks, mark=square*, mark options={mark size=1}] 
		table[x expr=\thisrow{x}-2000,y=ClusterHeadacheT,col sep=comma]
		{chapters/tables/JoinedResultProbsEPSamplesJTBPInferredWithInf.csv};
		\addplot[purple, only marks, mark=triangle*, mark options={mark size=1,rotate=180}] 
		table[x expr=\thisrow{x}-2000,y=TensionHeadacheT,col sep=comma] 
		{chapters/tables/JoinedResultProbsEPSamplesJTBPInferredWithInf.csv};
		\addplot[orange, only marks, mark=diamond*, mark options={mark size=1}] 
		table[x expr=\thisrow{x}-2000,y=MigraineWithoutAuraT,col sep=comma] 
		{chapters/tables/JoinedResultProbsEPSamplesJTBPInferredWithInf.csv};
		\addplot[green, only marks, mark=triangle*, mark options={mark size=1}] 
		table[x expr=\thisrow{x}-2000,y=MigraineWithAuraT,col sep=comma] 
		{chapters/tables/JoinedResultProbsEPSamplesJTBPInferredWithInf.csv};
		\addplot[red, only marks, mark=*, mark options={mark size=1}] 
		table[x expr=\thisrow{x}-2000,y=BrainTumorT,col sep=comma] 
		{chapters/tables/JoinedResultProbsEPSamplesJTBPInferredWithInf.csv};
		\legend{BrainTumor,MigraineWithAura,MigraineWithoutAura,TensionHeadache,ClusterHeadache}
		\end{axis}
		\end{tikzpicture}}
	\caption{Results of the inference with join tree belief propagation for the examples generated with EP.}\label{fig:results:full_plot:ep_samples:jtbp_inference}
\end{figure}
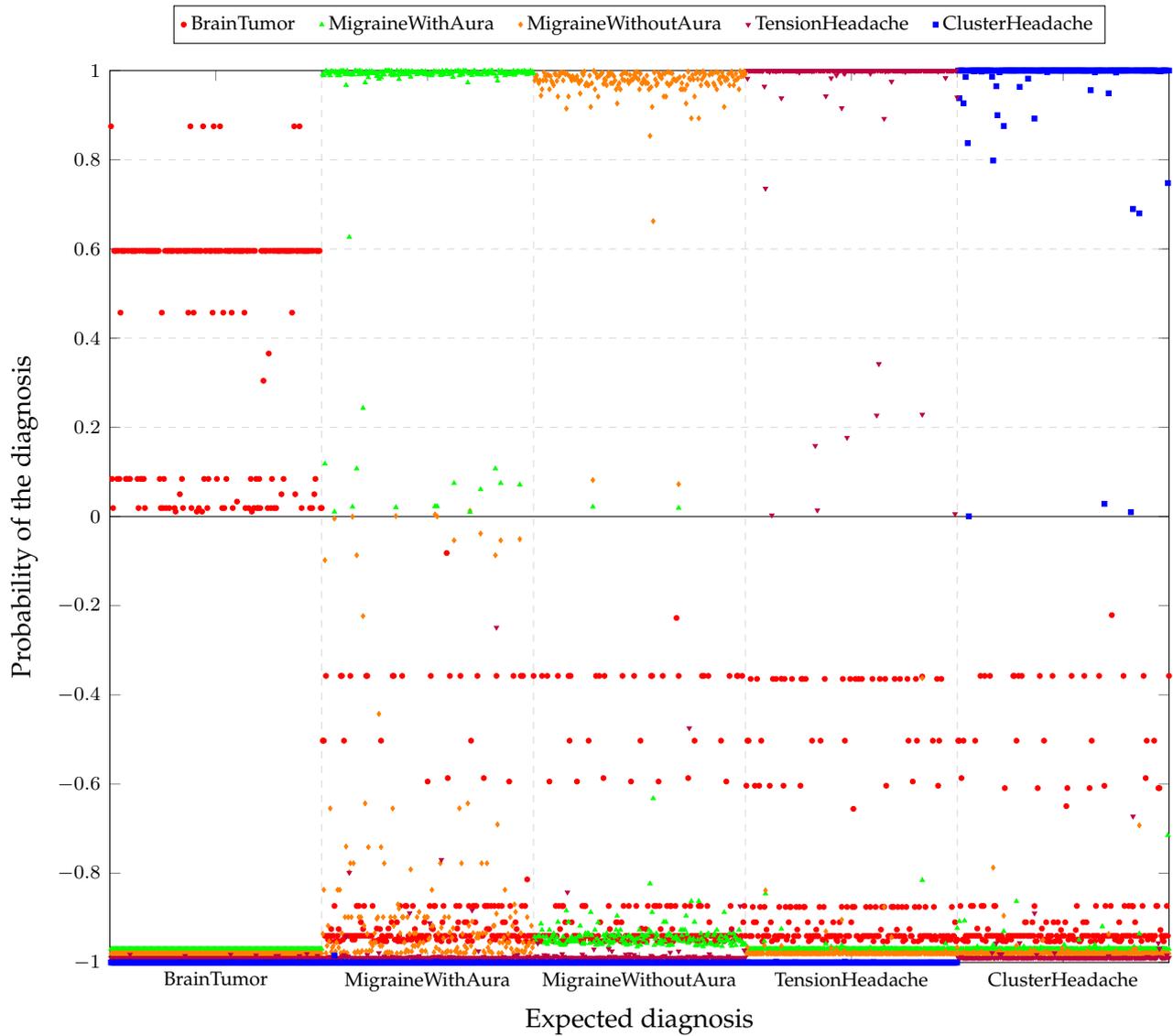


\subsection{Evaluation of the results}

\subsubsection{Results of the inference with junction tree belief propagation and expectation propagation}

Table~\ref{tab:results:correct_classifications} and figure~\ref{fig:results:full_plot:ep_samples:ep_inference} to~\ref{fig:results:full_plot:ep_samples:vmp_inference} show that the inferences with JTBP with the network from~\cite{beleg} and EP with the network from figure~\ref{fig:full_network} give very good results for all example data sets, regardless of the algorithm and network they were generated with. In addition, the results of JTBP and EP hardly differ from each other.
Furthermore, figures~\ref{fig:results:full_plot:ep_samples:ep_inference} to~\ref{fig:results:full_plot:ep_samples:vmp_inference} show that only the range of probabilities is slightly higher for JTBP than for EP\@. This may also be partly due to the other network used with this algorithm, but does not change the proportion of correctly classified examples (figure~\ref{fig:results:full_plot:ep_samples:ep_inference}).  The few misclassified examples are mostly accounted for by randomness during sample generation, where a characteristic that is critical in terms of differential diagnostics was generated with a value that does not support the expected diagnosis. For example, in wrongly classified examples where migraine without aura was expected, the \textit{aura symptoms} characteristic was generated with the value \textit{true}, and for expected migraines with aura with the value \textit{false}.
Besides, false classifications can be expected to occur particularly in examples where similar probabilities are calculated for two diagnoses. This case is very rare and only arises if all diagnoses have very low probabilities anyway. There are a few examples where tension headache or cluster headache is expected and brain tumour is the (relatively) most likely diagnosis, though with a very low absolute probability. Again, a typical brain tumour symptom was created with a positive value in the example generation, so that this value supports the wrong diagnosis.  

\subsubsection{Results of the inference with variational message passing}\label{chapter:implementation:sec:results:subsec:vmp}

Variational message passing shows similarly good results as EP and JTBP for four of the five diagnoses. However, in examples where \textit{migraine with aura} is expected  it is more likely to classify incorrectly than correctly (table~\ref{tab:results:correct_classifications}). A closer inspection of the symptoms indicates that the cause does not lie in the examples, since examples where the characteristic values support the expected diagnosis to a large extent are wrongly classified. After all, most of these examples could be correctly classified by EP and JTBP. 

\subsubsection{Conclusion}

Expectation propagation and join tree belief propagation provide very good results for networks in medical diagnostics. Although VMP yields comparable results for most diagnoses, a partial quality loss is unacceptable. According to the current state of knowledge, both EP and JTBP are suitable for carrying out queries on a multiply connected discrete Bayesian network.  Of the algorithms implemented in Infer.NET, expectation propagation is the method of choice for the intended application and clearly preferable to variational message passing.

\section{Summary}

In order to evaluate the quality of a Bayesian network and an inference algorithm, a prototypical software system was created to generate sample queries for and evaluate them with the help of the network. The software system was used to analyse the algorithms \textit{expectation propagation} and \textit{variational message passing} provided by Infer.NET and to compare them with \textit{join tree belief propagation}. Likewise, the Bayesian network from figure~\ref{fig:full_network} was built and found to provide very good results for the model domain of headache diagnostics. It was also shown that EP is more suitable as an inference algorithm for this domain than VMP and that it delivers an inference quality as high as with JTBP.

If, for example, the domain of pain diagnostics is to be covered completely, the construction of additional Bayesian networks is necessary. 
In~\cite{Nikovski_2000_3154_2} suitable construction methods which address these problems and can be applied in an iterative design process are described. For the determination of necessary a priori probabilities, distributions of classificatory characteristics can be obtained by the method~\cite{PDZ19}. 


In order to use the constructed Bayesian networks in diagnostic systems, the developed agent must be integrated into the respective system and connected to the existing interface for querying anamnesis data. The information from Bayesian networks can be used in a way that goes beyond the mere determination of diagnostic probabilities for a given anamnesis.



\bibliographystyle{abbrvdin}
\bibliography{diplom}

\end{document}